\documentclass[journal,twoside,web]{ieeetran}
\usepackage{generic}
\usepackage{cite}
\usepackage{amsmath,amssymb,amsfonts}
\usepackage{graphicx}
\usepackage{algorithm}
\usepackage{hyperref}
\hypersetup{hidelinks}
\usepackage{textcomp}
\usepackage{multirow}
\usepackage{mathrsfs}
\usepackage{manyfoot}
\usepackage{booktabs}
\usepackage{algorithmicx}
\usepackage{algpseudocode}
\usepackage{listings}
\usepackage[most]{tcolorbox}
\usepackage{xcolor}
\usepackage{fvextra}
\usepackage{array}
\usepackage{tabularx}

\usepackage{amsthm}

\newtcolorbox{promptbox}{
  width=\columnwidth, 
  colback= gray!5,
  colframe=gray!70!black,
  boxrule=0.8pt,
  arc=2mm,
  left=3mm, right=3mm, top=2mm, bottom=2mm,
  center,
  enhanced,
  breakable,
}

\def\BibTeX{{\rm B\kern-.05em{\sc i\kern-.025em b}\kern-.08em
    T\kern-.1667em\lower.7ex\hbox{E}\kern-.125emX}}
\begin{document}
\title{Voice-Interactive Surgical Agent for \\ Multimodal Patient Data Control}
\author{Hyeryun Park, Byung Mo Gu, Jun Hee Lee, Byeong Hyeon Choi, Sekeun Kim, Hyun Koo Kim, and Kyungsang Kim
\thanks{
This research was supported by a grant of the Korea Health Technology R\&D Project through the Korea Health Industry Development Institute (KHIDI), funded by the Ministry of Health \& Welfare, Republic of Korea (grant number: RS-2024-00436472).
Additionally, this work was supported by the National Research Foundation of Korea (NRF) grant, funded by the Korea government (MSIT) (grant number: RS-2025-00518091).
(Corresponding authors: Hyun Koo Kim; Kyungsang Kim.) All authors declare no conflict of interest.}
\thanks{H. Park is with the Image Guided Precision Cancer Surgery Institute, College of Medicine, Korea University, Seoul, Republic of Korea, and also with the Department of Radiology, Massachusetts General Hospital, Boston, Massachusetts, USA (e-mail: hyerpark1115@gmail.com).}
\thanks{B. M. Gu and J. H. Lee are with the Image Guided Precision Cancer Surgery Institute, College of Medicine, Korea University, Seoul, Republic of Korea, and the Department of Thoracic and Cardiovascular Surgery, Korea University Guro Hospital, College of Medicine, Korea University, Seoul, Republic of Korea (e-mail: luvotomy7@naver.com; lee2632@naver.com).}
\thanks{B. H. Choi and H. K. Kim are with the Image Guided Precision Cancer Surgery Institute, College of Medicine, Korea University, Seoul, Republic of Korea, and the Department of Thoracic and Cardiovascular Surgery, Korea University Guro Hospital, College of Medicine, Korea University, Seoul, Republic of Korea, and the Department of Biomedical Sciences, College of Medicine, Korea University, Seoul, Republic of Korea (e-mail: baby2music@gmail.com; kimhyunkoo@korea.ac.kr).}
\thanks{S. Kim and K. Kim are with the Department of Radiology, Massachusetts General Hospital and Harvard Medical School, Boston, Massachusetts, USA (e-mail: skim207@mgh.harvard.edu; kkim24@mgb.org).}}

\maketitle

\begin{abstract}

In robotic surgery, surgeons fully engage their hands and visual attention in procedures, making it difficult to access and manipulate multimodal patient data without interrupting the workflow.
To overcome this problem, we propose a Voice-Interactive Surgical Agent (VISA) built on a hierarchical multi-agent framework consisting of an orchestration agent and three task-specific agents driven by Large Language Models (LLMs).
These LLM-based agents autonomously plan, refine, validate, and reason to interpret voice commands and execute tasks such as retrieving clinical information, manipulating CT scans, or navigating 3D anatomical models within surgical video.
We construct a dataset of 240 user commands organized into hierarchical categories and introduce the Multi-level Orchestration Evaluation Metric (MOEM) that evaluates the performance and robustness at both the command and category levels.
Experimental results demonstrate that VISA achieves high stage-level accuracy and workflow-level success rates, while also enhancing its robustness by correcting transcription errors, resolving linguistic ambiguity, and interpreting diverse free-form expressions.
These findings highlight the strong potential of VISA to support robotic surgery and its scalability for integrating new functions and agents.
\end{abstract}


\section{Introduction}
\label{sec:introduction}
\IEEEPARstart{T}{he} primary surgeon must maintain concentration, with both hands and eyes fully engaged throughout robotic surgery.
However, accessing and manipulating patient data during the procedure requires shifting attention away from the surgeon console to another monitor or auxiliary interfaces.
These interruptions become more significant when surgeons need to check diverse multimodal data such as clinical information, CT images, MRI images, and 3D models, underscoring the need for a dynamic interaction system.
Early interaction studies on voice-controlled robotic surgery mainly focused on camera and equipment manipulation using systems such as AESOP \cite{allaf1998laparoscopic, mettler1998one, reichenspurner1999use, nathan2006voice}, the HERMES operating room control center \cite{salama2005utility}, and PARAMIS \cite{vaida2010development}.
These systems reduced operation time and the need for a second assistant \cite{nathan2006voice}, while improving surgical efficiency and decreasing communication burden \cite{salama2005utility}.
Subsequent studies broadened the scope of tasks, including image retrieval and 3D distance measurement \cite{forte2022design}, real-time camera control \cite{elazzazi2022natural, kim2025speech}, and tissue manipulation with robotic assistant arms \cite{davila2024voice}.
Recent studies have progressed toward interpreting natural language commands utilizing Large Language Models (LLMs).
A ChatGPT-integrated system mapped free-form user speech to the most relevant predefined commands \cite{pandya2023chatgpt}.
LLM-based mixed reality systems further enable natural language–driven robot operation \cite{zhang2025llm} and real-time surgical assessment and procedural guidance \cite{gavric2025surgery}.

LLMs exhibit advanced capabilities for understanding and generating natural language and such models include GPT-4 \cite{Touvron2023}, PaLM \cite{Chowdhery2023}, LLaMA \cite{Touvron:2023:LLaMA, Touvron:2023:LLaMA2, Dubey:2024:LLaMA3, Meta:2025:LLaMA4}, and Gemma \cite{Gemma:2024, Gemma2:2024, Gemma3:2025}.
These models can interpret natural instructions and perform multi-step problem solving through advanced prompting and reasoning techniques, including Chain-of-Thought (CoT) \cite{Wei2022}, Self-Consistency CoT \cite{wang2022scCoT}, Tree-of-Thought \cite{yao2023ToT}, and Graph-of-Thought \cite{besta2024GoT}.
These advancements in reasoning have transformed LLMs from language generators into autonomous agents that can perceive their environment, plan to achieve specific goals, and act accordingly \cite{franklin1996agent, russell1995modern, huang2025introduction}.
Various frameworks such as ReAct \cite{yao2022react}, AutoGPT, BabyAGI, AutoGen \cite{wu2024autogen}, and CAMEL \cite{li2023camel} have been proposed to integrate reasoning and action.
Recent research further applies reinforcement learning to help LLMs learn from past reasoning failures and optimize their decision-making \cite{zhang2025survey}.
These diverse reasoning and interaction methods have become the driving force of LLM-based autonomous agents capable of planning, tool use, and collaborative problem solving.

\begin{figure*}[t]
    \centering
    \includegraphics[width=0.9\textwidth]{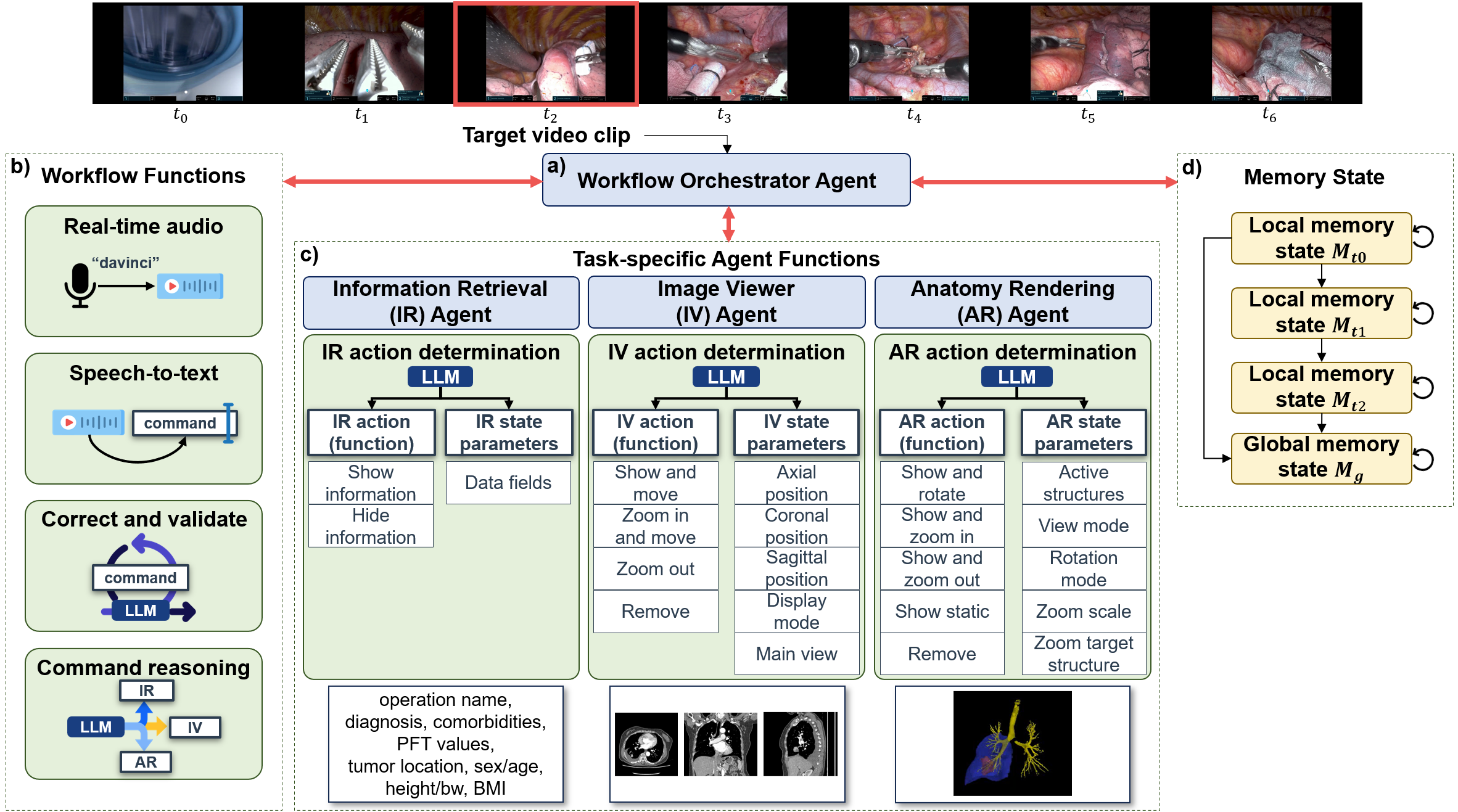}
    \caption{Overall framework of the proposed VISA: (a) Workflow orchestrator agent that autonomously plans the order of function execution; (b) Workflow functions that capture, transcribe, correct, and route voice commands; (c) Task-specific agent functions that control patient data on the target video clip; (d) Memory state with local memory for each clip and a global memory for shared contextual understanding across clips.}
    \label{VISA_workflow}
\end{figure*} 

The LLM-based agent typically consists of five key modules: a profile module that defines objectives and constraints, a memory module that stores past states, a reasoning and planning module that formulates strategies, an action module that executes tasks via models or tools, and a feedback module that evaluates results and refines behavior \cite{wang2024survey}.
These agents can operate as a single-agent system using one LLM or as a multi-agent system (MAS) with specialized agents for distinct tasks, offering advantages in scalability, efficiency, and coordinated workflows \cite{li2024survey, zhao2025llm}.
The MAS coordination is enabled by LLM-based orchestration, where an LLM functions as the cognitive core of the system to manage interactions among agents, plan execution sequences, interpret intermediate results, and control the overall workflow \cite{tran2025multi, rasal2024navigating, du2025multi, dang2025multi, zhang2025osc, zhao2025llm, tran2025multi}.
Such systems may adopt centralized, decentralized, or hierarchical structures and exhibit cooperative, competitive, or negotiated behaviors \cite{zhao2025llm, tran2025multi}.
To implement practical orchestration, various agent frameworks have emerged such as LangGraph \cite{LangGraph2024}, AutoGen \cite{wu2024autogen}, LlamaIndex \cite{Liu_LlamaIndex_2022}, CrewAI, and AgentKit.

In this work, we highlight the importance of effective surgical tool interaction during operation, with voice interaction offering a promising approach.
Based on this motivation, we modularize key tool functionalities into specialized agents and coordinate them through an orchestrator. 
Specifically, we introduce a Voice-Interactive Surgical Agent (VISA) that interprets free-form voice commands and enables context-aware multimodal data interaction directly on surgical videos.
To our knowledge, this is the first hierarchical multi-agent framework, consisting of a workflow orchestrator agent and multiple task-specific agents as demonstrated in Fig.\ref{VISA_workflow}.

In contrast to prior studies that rely on predefined commands or a single LLM for direct voice-to-function mapping, our VISA autonomously plans and executes functions through intermediate steps, providing more flexible handling of complex voice commands.
The workflow orchestrator agent uses an LLM to perform probabilistic decision-making across workflow functions and task-specific agent functions, predicting the next function to execute.
The workflow functions capture audio, perform Speech-to-Text (STT), correct and validate commands, and select the appropriate task agent.
The Information Retrieval (IR) agent retrieves clinical information, the Image Viewer (IV) agent slides through CT scans, and the Anatomy Rendering (AR) agent manipulates 3D anatomical models.
Each task-specific agent includes an action determination function that leverages an LLM to infer the appropriate action and its parameters.
This hierarchical design enables scalable integration of new functions or agents. 
VISA also incorporates a memory state that stores previous commands, agents, and various parameters for continuous task execution and adaptive responses to ambiguous commands.

For evaluation, we construct a command dataset of 240 user commands, each annotated with three hierarchical categories of command structure, type, and expression.
We also introduce a Multi-level Orchestration Evaluation Metric (MOEM) to assess stage-level accuracy and workflow-level success rates.
Experimental results demonstrate that VISA achieves high performance and strong robustness to linguistic variability, speech recognition errors, and ambiguous commands through LLM-based command correction and reasoning.
Further analyses considering real surgical settings and comparisons with synthesized speech underscore the strong potential of VISA.

\begin{figure*}[t]
    \centering
    \includegraphics[width=0.9\textwidth]{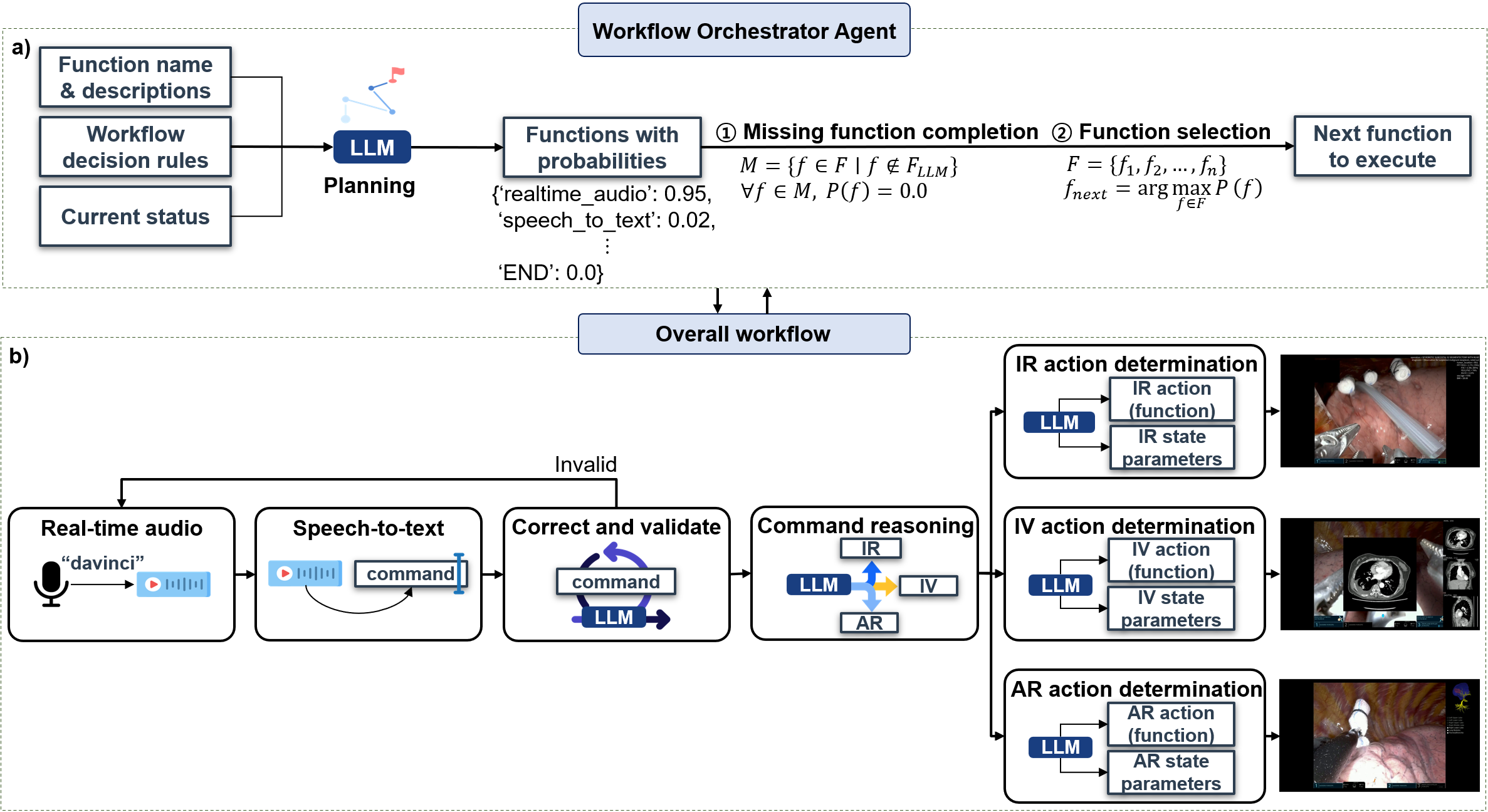}
    \caption{Overall workflow planned by the workflow orchestrator agent with a hybrid approach of LLM and mathematical rules.}
    \label{LLMOrchestrator_and_workflow}
\end{figure*} 

\section{Methods}

\subsection{Overall Workflow}

Our VISA framework is a hierarchical MAS that automatically plans and executes the overall workflow as shown in Fig.~\ref{LLMOrchestrator_and_workflow}.
We divide the surgical video into 10-second clips, with each clip designed to handle a single command.
The workflow orchestrator agent initializes the memory state for the target video clip and uses the LLM to select the next function to execute.
The LLM prompt includes instructions, function names with definitions, predifined workflow decision rules, the current status, and the output format.

In determining return points or termination, relying solely on function names and descriptions can lead to mispredictions.
To mitigate this, workflow decision rules guide the overall process and define when to return and end.
In addition, current status such as “No audio recorded”, “Last command invalid, need new input”, and “Agent completed, workflow finished” provide contextual cues that help navigate branching points.
The workflow decision rules and current status allow the orchestrator to make more accurate and consistent decisions in various scenarios.

The LLM output should be in json format, containing each function name along with its associated probability for the next step. 
Incorporating probabilities allows the system to provide a quantitative measure of confidence for decision-making.
Since the LLM output may omit certain functions, the missing function completion rule includes all absent functions $M$ with their probabilities set to zero.
Subsequently, the function selection rule chooses the next function $f_{next}$ with the highest probability over the complete function set.
This hybrid approach combines the capabilities of LLMs with the probability-based decision rules, ensuring reliability and mathematical grounding.

\subsection{Workflow Functions}

\subsubsection{Real-time Audio Function}

This function handles voice interaction by capturing audio on the edge device and transmitting it to a remote server.
For fast real-time processing on CPU, it integrates Silero-VAD\footnote{\url{https://github.com/snakers4/silero-vad}} for voice activity detection (VAD), Whisper-small model \cite{pmlr-v202-radford23a} via faster-whisper\footnote{\url{https://github.com/SYSTRAN/faster-whisper}} for speech recognition and wake-word detection, and Microsoft Edge TTS\footnote{\url{https://github.com/rany2/edge-tts}} for text-to-speech (TTS).
The process begins with microphone streaming, where VAD detects speech activity, Whisper transcribes the audio to identify the wake-word "davinci", records the command until four seconds of silence, and sends it to the remote server.
Wake-word detection is further improved by optimizing Whisper-small decoding parameters, setting “davinci” as a hotword, the temperature to $0$, beam size to $5$, no-speech threshold to $0.25$, compression ratio threshold to $1.2$, and log probability-threshold to $-1.0$.

\subsubsection{Speech-to-Text Function}
The STT function employs the Whisper-medium model \cite{pmlr-v202-radford23a} via faster-whisper for efficient inference.
Among the SpeechBrain model \cite{ravanelli2021speechbrain}, the Whisper models, and Microsoft Phi-4-multimodal model \cite{microsoft2025phi4}, the Whisper-medium model offers a practical balance between speed and accuracy.
Testing on 35 randomly selected commands, larger models generally achieve higher accuracy but incur slower inference.
The Whisper-medium model is configured with a beam size of $8$, the language set to “en”, a temperature of $0$, a speech threshold of $0.3$, and a log prob threshold of $-2.0$ to enhance transcription accuracy for English speech while effectively filtering out non-speech segments and low-confidence outputs.
Lastly, the transcribed command is updated in the memory state.

\subsubsection{Correct and Validate Function}
This function employs Gemma3 \cite{Gemma3:2025} LLM to correct the transcribed command and determine its validity.
The prompt\footnote{\label{prompt}Prompts available at \url{https://github.com/helena-lena/SAOP}} includes the available task agents with their descriptions, instructions, the current transcribed command, correction rules, validation rules, the output format, and the global memory state.
Since the STT model is unfamiliar with medical terminology, we provide correction rules such as replacing “city” with “CT” or COVID-related terms with “coronal”.
Ambiguity also arises with commands like “zoom in/out”, which may refer either to the IV agent or the AR agent.
To resolve these problems, we provide the last three revised commands and associated agents from the global memory state.
If the command is “None”, the LLM revises it to “Select \{agent name\}” by inheriting the agent from the previous memory state.
The LLM outputs a json containing the corrected command and its validity, which is updated in the memory state.
Valid commands proceed to the command reasoning function, while invalid ones return to the real-time audio function for a new voice input.

\subsubsection{Command Reasoning Function}
This function takes the revised command as input and employs Gemma3 model to reason and select the most appropriate task-specific agent.
The prompt\footnotemark[\getrefnumber{prompt}] includes the available task agents with brief descriptions, instructions, the current revised command, and global memory state.
Providing the last three revised commands and their corresponding agents from the global memory state helps the model resolve ambiguous commands.
We also apply Chain-of-Thought (CoT) prompting \cite{Wei2022} to the reasoning process to analyze the cause of failures and use these insights to refine agent descriptions.
The model finally returns a json specifying the name of the selected agent, which is also updated in the memory state.

\subsection{Task-specific Agent Functions}

\subsubsection{IR Agent Functions}

\begin{figure}[b]
    \centering
    \includegraphics[width=\linewidth]{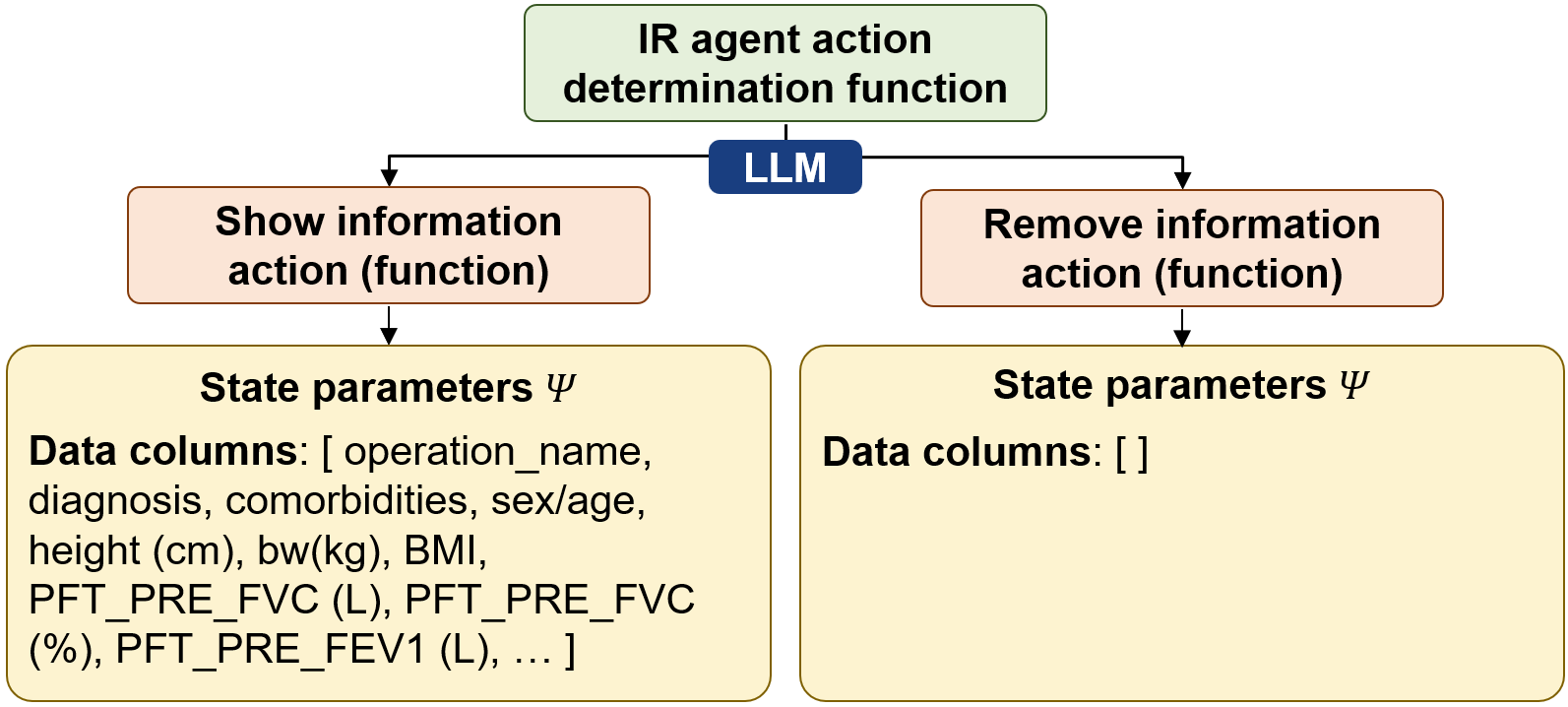}
    \caption{The overview of the IR agent, which executes either the show or remove function with inferred state parameters.}
    \label{patient_info_agent} 
\end{figure}

The IR agent overlays the requested clinical data on the surgical video and consists of three functions: an action determiner (AD) function that interprets commands and decides the action and parameters, and two functions for showing or removing the overlay as shown in Fig.~\ref{patient_info_agent}.
The AD function leverages Gemma3 to infer the action uniquely mapped to an agent function and state parameters\footnotemark[\getrefnumber{prompt}].
We denote the command by $x \in \mathcal{X}$, the instruction by $i$, and a set of patient data columns by $\mathcal{C}=\{c_1,\dots,c_m\}$ with corresponding values $v=\{v_1,\dots,v_m\}$.
The LLM parameterized by $\theta$ produces an action distribution $\pi_\theta$ over $\mathcal{A}=\{\text{SHOW},\text{HIDE}\}$ as in (\ref{eq:pt_action}) and a per-field inclusion probability $\phi_\theta$ as in (\ref{eq:pt_info_eq1}):
\begin{align}
&\pi_\theta(a \mid x,i) = \Pr(\mathcal{A}=a \mid x,i), \label{eq:pt_action} \\
&\phi_\theta(j \mid x,i,a) = \Pr(Y_j=1 \mid x,i,a), \quad j=1,\dots,m. \label{eq:pt_info_eq1}
\end{align}
The agent selects the action $a^\star$ with the higher probability and predicts the indicator vector $y^\star \in \{0,1\}^m$ that specifies which fields to display, as in (\ref{pt_info_eq2}). 
To handle ambiguous expressions like physical information, the prompt guides their mapping to specific fields in $\mathcal{C}$, adjusting the $\phi_\theta$.
\begin{align}
a^\star = \arg\max_{a \in \{\text{SHOW},\text{HIDE}\}} \pi_\theta(a \mid x,i), \\
y_j^\star = \mathbf{1}\!\left[\phi_\theta(j \mid x,i,a^\star) \ge \tau \right], \label{pt_info_eq2}
\quad j=1,\dots,m. 
\end{align}

The state parameters of the IR agent are defined as:
\begin{equation}
\Psi = (y, s), \quad
\begin{aligned}
  &y \in \{0,1\}^m,
  &s \in \text{final string}.
\end{aligned}
\end{equation}
Here, $y$ denotes the binary indicator vector specifying which data fields are active, and $s$ is the final patient information string to overlay.  
When $a^\star=\text{HIDE}$, the agent sets $y^\star=\mathbf{0}$ and assigns an empty string $s$.  
When $a^\star=\text{SHOW}$, the agent retrieves the relevant clinical data and formats it according to physician preferences to construct $s$.
Given the indicator vector $y^\star$, the subset of columns to display is:
\begin{align}
\mathcal{C}_{y^\star} = \{\, c_j \in \mathcal{C} \mid y_j^\star = 1 \,\}.
\label{eq:colset}
\end{align}
For each selected column $c_j \in \mathcal{C}_{y^\star}$ with value $v_j$, a field-specific formatting function $r(c_j,v_j)$ transforms the raw data into a textual representation as in (\ref{eq:formatter}).
\begin{align}
r(c_j, v_j) : (c_j, v_j) \mapsto \text{string},
\quad c_j \in \mathcal{C}_{y^\star}.
\label{eq:formatter}
\end{align}
This formatting function reflects physician feedback, such as combining pulmonary function test values in liters and percentages, merging relevant fields like sex and age, and omitting unnecessary units. 
The final patient information string is the result of concatenating all the formatted outputs with newline delimiters as in (\ref{eq:string_comp}).
\begin{align}
s = \mathrm{concat}_{c_j \in \mathcal{C}_{y^\star}} \; r(c_j,v_j), \text{delimiter = ``\textbackslash n''}.
\label{eq:string_comp}
\end{align}

Finally, the SHOW action links to a display function that overlays the generated text on the top-right corner of each video frame. 
In contrast, the HIDE action connects to a removal function that restores the original frame without any overlay.  
Given a sequence of video frames $\{I_t\}_{t=1}^T$ and an overlay box position $p$, the overlay operator $\mathcal{O}_{IR}$ places the patient information string $s$ on each frame as follows:
\begin{align}
\tilde I_t = \mathcal{O}_{IR}(I_t; s, p), \quad t=1,\dots,T. \label{eq:overlay}
\end{align}
The selection of data columns and the visual presentation style is refined through continuous consultation with clinicians to ensure readability.

\begin{figure*}[t]
    \centering
    \includegraphics[width=\textwidth]{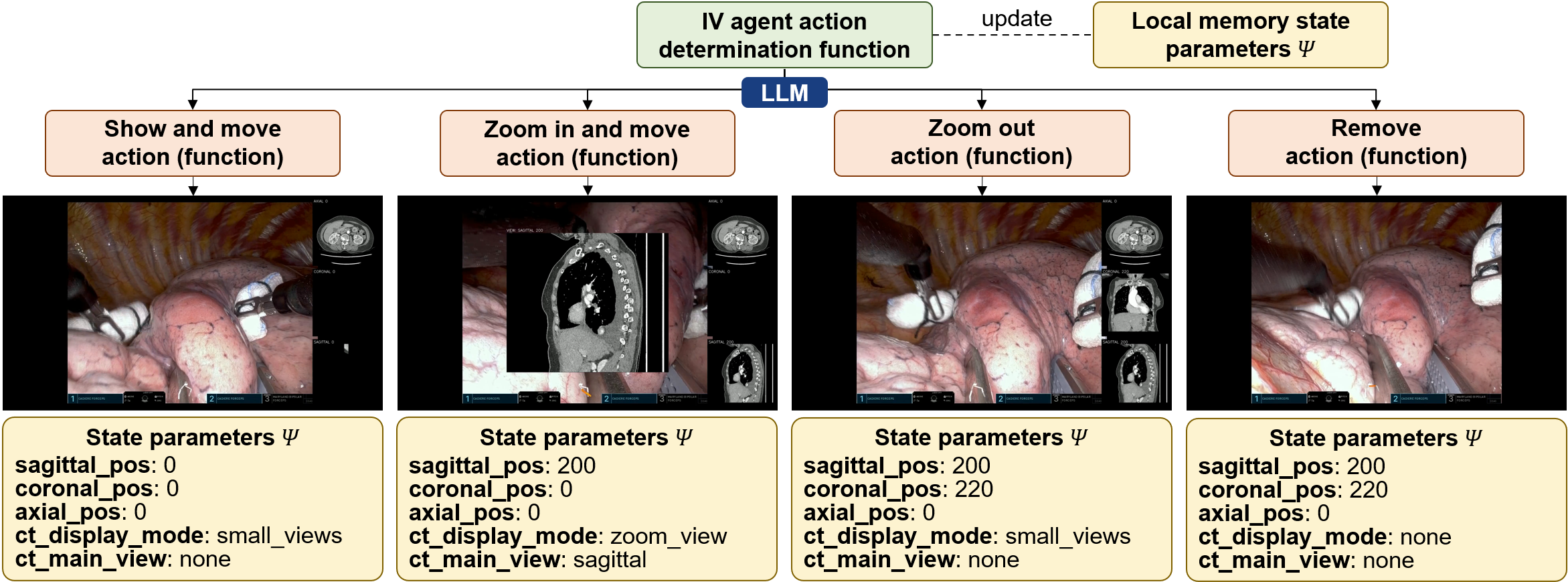}
    \caption{The overview of the IV agent, which executes show, zoom in, zoom out, or remove function with inferred state parameters.}
    \label{CT_agent} 
\end{figure*}  

\subsubsection{IV Agent Functions}

The IV agent overlays CT DICOM images onto the surgical video, allowing surgeons to scroll through axial, coronal, and sagittal planes as shown in Fig.~\ref{CT_agent}.
The AD function interprets commands and selects one of four actions: displaying and moving multi-planar views, zooming into a large single view, zooming out to the small multi-planar views, and removing all CT views\footnotemark[\getrefnumber{prompt}].
Given the command $x \in \mathcal{X}$ and the instructions $i$, the Gemma3 model parameterized by $\theta$ infers the action $a \in \mathcal{A}$ and updates the state parameters $\Psi$ following the joint policy:
\begin{align}
\pi_\theta(a,\Psi \mid x,i)
&= \pi_\theta(a \mid x,i)\;\pi_\theta(\Psi \mid x,i,a),  \\
a^\star &= \arg\max_{a \in \mathcal{A}} \ \pi_\theta(a \mid x,i),  \\
\Psi^\star &= \arg\max_{\Psi} \ \pi_\theta(\Psi \mid x,i,a^\star). \label{eq:fact_state_argmax}
\end{align}
The policy first infers the action $a^\star$, and then predicts the updated state parameters $\Psi^\star$ conditioned on that action.
The $\Psi$ denotes the current state and $\Psi^\star$ is the updated state.

Specifically, the state parameters are defined as $\Psi = (p, \delta, v)$, where
$p = (p^{\mathrm{axi}}, p^{\mathrm{cor}}, p^{\mathrm{sag}})$ is the slice positions,
$\delta \in \{\text{none}, \text{small\_views}, \text{zoom\_view}\}$ is the display mode,
and $v \in \{\text{axial}, \text{coronal}, \text{sagittal}\}$ is the main view plane.
The slice positions are updated based on movement commands:
\begin{align}
p^\star = p + \Delta p(x,i,a^\star),
\label{eq:ct_position}
\end{align}
where $\Delta p$ encodes positional adjustments such as ``move right'' or ``sagittal plus 50'' (increase sagittal), ``move down'' or ``axial minus 100'' (decrease axial), and ``move forward'' or ``coronal plus 30'' (increase coronal). 
The display mode $\delta$ determines how CT slices appear on the surgical video.
If $a^\star$ is $\text{SHOW\_MOVE}$, $\delta$ is small\_views, showing three CT views $I_t^{\mathrm{axi}}, I_t^{\mathrm{cor}}, I_t^{\mathrm{sag}}$ on the right side of the video, each with the updated positions $p^\star$.
If $a^\star$ is $\text{ZOOM\_IN\_MOVE}$, $\delta$ is zoom\_view, displaying a main CT view $I_t^{\mathrm{main}}$ in the center of the video.
The main view $v$ is either explicitly specified in the command or implicitly inferred from the movement axis: $I_t^{\mathrm{main}} = \mathcal{V}(p^\star, v)$.
If $a^\star=\text{ZOOM\_OUT}$, $\delta$ is \text{small\_views}, removing the large main CT view.
If $a^\star=\text{REMOVE}$, $\delta$ is \text{none}, which removes all CT views.

Given video frames $\{I_t\}_{t=1}^T$, action $a^\star$, and the updated state $\Psi^\star$, the overlay operator generates
\begin{align}
\tilde I_t = \mathcal{O}_{IV}(I_t; p^\star, a^\star, \delta, v), \quad t=1,\dots,T.
\label{eq:ct_overlay}
\end{align}
To ensure smooth transitions, $\mathcal{O}_{IV}$ interpolates slice positions from $p$ to $p^\star$ over five seconds.

\begin{figure*}[t]
    \centering
    \includegraphics[width=\textwidth]{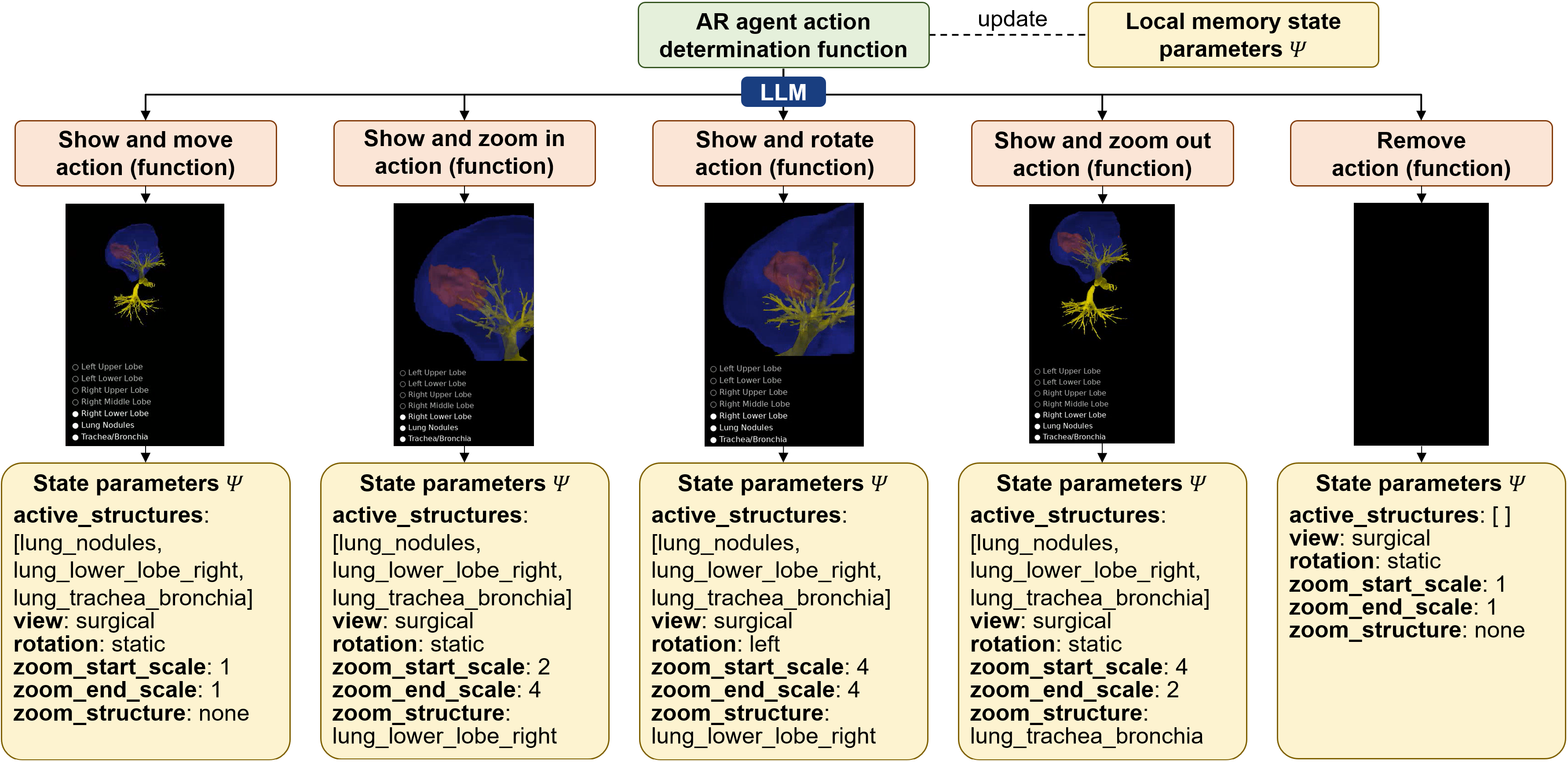}
    \caption{The overview of the AR agent. The action determination function predicts the action and state parameters, then the AR agent executes static, zoom in, rotate, zoom out, or remove function with predicted parameters.}
    \label{Recon_agent} 
\end{figure*}  

\subsubsection{AR Agent Functions}

The AR agent renders 3D anatomical models onto the surgical video for interactive visualization as shown in Fig.~\ref{Recon_agent}. 
The 3D models are reconstructed from CT images using TotalSegmentator\footnote{\url{https://github.com/wasserth/TotalSegmentator}} in the 3D Slicer\footnote{\url{https://www.slicer.org/}}, which generates anatomical structures with corresponding labels.
The AD function interprets commands and selects one of five actions: displaying static views, rotating the model, zooming in on specific structures, zooming out, and removing all models\footnotemark[\getrefnumber{prompt}]. 
The Gemma3 LLM jointly infers the action $a$ and updates the state parameters $\Psi$ following the same joint policy formulation as (\ref{eq:fact_state_argmax}) in the IV agent.
The difference is that $a$, $\Psi$, and $i$ are all specific to the AR Agent.
The state parameters $\Psi$ include active anatomical structures $\alpha$, viewpoint $v$, rotation mode $r$, target structure for zoom $\tau$, and zoom parameters $z$.

The active structures $\alpha$ represent the list of visible anatomical labels.
The physician selects the labels of interest, including left lower lobe, left upper lobe, right lower lobe, right middle lobe, right upper lobe, lung nodules, and the trachea/bronchia. 
The default setting activates the lobe containing the nodules together with all non-lobe structures, and subsequent commands add or remove structures accordingly.

The viewpoint $v \in$ \{\text{anterior}, \text{posterior}, \text{left}, \text{right}, \text{superior}, \text{inferior}, \text{surgical}\} defines the initial view of the 3D model.
The default viewpoint is set to surgical view, which aligns to the perspective of a patient lying on the operating table as shown in Figure~\ref{Recon_agent}.
For example, commands like ``Show the anterior view'' or ``Show from the front'' display the anterior side, while rotation commands such as ``Rotate the posterior view to the left'' set the $v$ to the posterior view.

The rotation mode $r \in$ \{\text{static}, \text{left}, \text{right}, \text{up}, \text{down}, \text{horizontal}, \text{vertical}\} is the relative rotation direction from the initial viewpoint $v$, with the static mode used by default.
The static mode leaves the model fixed, the left/right/up/down modes rotate by $30$ degrees toward the specified direction and return, and the horizontal/vertical modes perform a continuous $360$ degrees rotation.

If $a^\star$ is zoom in/out, the LLM extracts the target structure $\tau$ from the command and updates the zoom parameters $z^\star = (z_c^\star, z_s^\star, z_\ell^\star)$, where $z_c^\star$ is the center of $\tau$, $z_s^\star$ is the zoom scale, and $z_\ell^\star$ is the zoom depth level.
The zoom in moves toward $z_c^\star$ and stores the zoom path in a history stack, which is later traversed in reverse order during the zoom out.
The zoom scale starts at $z_s = 1.0$, doubles when zooming in, and halves when zooming out until reaching the minimum scale of $1.0$, while the zoom depth level increases or decreases accordingly.

Given video frames $\{I_t\}_{t=1}^T$, action $a$, and updated state $\Psi^\star$, the overlay operator $\mathcal{O}_{\text{AR}}$ generates
\begin{align}
\tilde I_t = \mathcal{O}_{\text{AR}}(I_t; \Psi^\star, a^\star), \quad t=1,\dots,T.
\label{eq:recon_overlay}
\end{align}
$\mathcal{O}_{\text{AR}}$ smoothly interpolates the rotation and zoom transitions, with zooms completing in three seconds, left/right/up/down rotations in seven seconds (three seconds movement, one second hold, three seconds return), and full horizontal/vertical rotations in six seconds.
The resulting overlays appear in the upper-right corner of the surgical video, allowing the surgeon to identify critical anatomical structures during the procedures.

\subsection{Dataset}
We construct a dataset of 240 user commands\footnote{\label{data_code}Data and code on \url{https://github.com/helena-lena/SAOP}}: 44 for the IR agent, 81 for the IV agent, and 115 for the AR agent. 
This distribution reflects the differing functional loads of the agents, with the AR agent accounting for nearly half of all commands due to its broader range of actions and parameters.
Each command also includes category annotations along three hierarchical dimensions: command structure, command type, and command expression, as summarized in Table~\ref{tab:command_category_distribution}.

The command structure dimension categorizes each command as single or composite, indicating whether it requires decomposition into sequential sub-commands.
The command type dimension classifies commands into explicit, implicit, and Natural Language Question (NLQ) forms, capturing different levels of linguistic clarity and ambiguity.
Finally, the command expression dimension groups commands into baseline, abbreviation, and paraphrase to reflect lexical and phrasing variability, as clinicians express the same intent using different terminology or styles.

\begin{table*}[t]
\centering
\caption{\textbf{Hierarchical command category definitions and distributions.}}
\label{tab:command_category_distribution}
\renewcommand{\arraystretch}{1.25}
\begin{tabular}{cclc}
\hline
\textbf{Dimension} & \textbf{Category} & \textbf{Description and Examples} & \textbf{Count} \\
\hline

\multirow{4}{*}{\shortstack{\textbf{Command}\\\textbf{Structure}}}
& Single & Definition: Contains one clearly defined action or data request. & 225\\[-1pt]
&        & Example: ``Show patient information.'', ``Display CT images.'', ``Remove models.'' & \\[3pt]
& Composite & Definition: Combines two or more actions or data requests. & 15 \\[-1pt]
&          & Example: ``Move axial to the middle slice and zoom in.'', ``Show physical and PFT info.'' & \\[3pt]
\hline

\multirow{6}{*}{\shortstack{\textbf{Command}\\\textbf{Type}}}
& Explicit & Definition: States both subject (data/agent) and verb (action), and does not require contextual reasoning. & 80 \\[-1pt]
&         & Example: ``Show the anterior view.'', ``Add the right upper lobe.'' & \\[3pt]
& Implicit & Definition: Omits either the subject or the verb, making it dependent on the previous memory context. & 80 \\[-1pt]
&         & Example: ``Zoom in.'', ``Initialize.'', ``Surgical view.'', ``Diagnosis.'', ``Rotate to the right.'' & \\[3pt]
& NLQ & Definition: Uses conversational or question-style phrasing. & 80 \\[-1pt]
&     & Example: ``How old is the patient?'', ``Can you activate the right lung?'' & \\[3pt]
\hline

\multirow{6}{*}{\shortstack{\textbf{Command}\\\textbf{Expression}}}
& Baseline & Definition: Expressions without any abbreviations or paraphrasing, matching phrasing defined in the prompt. & 145 \\[-1pt]
&          & Example: ``Display the pulmonary function test.'', ``Coronal plus 256.'', ``Show from the anterior.'' & \\[3pt]
& Abbreviation & Definition: Includes abbreviated forms used in both medical terminology and general expressions. & 15 \\[-1pt]
&              & Example: ``Display the PFT info.'', ``Remove RUL.'', ``Show patient info.'' & \\[3pt]
& Paraphrase & Definition: Expresses the same intent as baseline but uses alternative synonyms or phrasing styles. & 80 \\[-1pt]
&            & Example: ``Move front, front, front.'', ``Move coronal to the middle slice.'', ``Show from the front.'' & \\[3pt]
\hline

\end{tabular}
\end{table*}

We additionally synthesize voice command datasets\footnotemark[\getrefnumber{data_code}] for four different speakers using Microsoft Edge TTS that we use in real-time speech function.
Among the available voices, we select two female and two male speakers categorized under “News”, as they are native English speakers and provide clear pronunciation.
Specifically, \text{en-US-AriaNeural} conveys polite and confident tones, \text{en-US-JennyNeural} sounds friendly, considerate, and comforting, \text{en-US-GuyNeural} expresses passion, and \text{en-US-ChristopherNeural} reflects reliability and authority.

Patient data were obtained from the Department of Thoracic and Cardiovascular Surgery at Korea University Guro Hospital and include surgical videos, clinical information, CT images, and 3D anatomical models from a patient who underwent lung surgery using the da Vinci robotic system.
This study was ethically approved by the Institutional Review Board at Korea University Guro Hospital (IRB No. 2025GR0018), and informed consent was waived due to its retrospective design.

\section{Evaluation Metrics}\label{results_evaluation_metrics_section}

For each command, we evaluate the outcome of each workflow stage using a binary scheme (1: correct, 0: incorrect), as shown in Table~\ref{command_results}.
STT checks whether the transcribed command, command correction (CC) evaluates the revised command, and command reasoning (CR) determines whether the LLM assigns the command to the appropriate agent.
Agent function (AF) and agent parameters (AP) are outputs of AD function, which is correct only if both AF and AP are accurate.
Orchestration flow (OF) assesses whether the workflow proceeds in the correct order, while invalid cycle (IC) is the number of times the workflow enters an invalid loop.
Finally, we introduce the Multi-level Orchestration Evaluation Metric (MOEM), which comprehensively evaluates VISA from two perspectives: command-level orchestration performance and command-category performance.

\begin{figure}[h]
    \centering
    \includegraphics[width=0.9\linewidth]{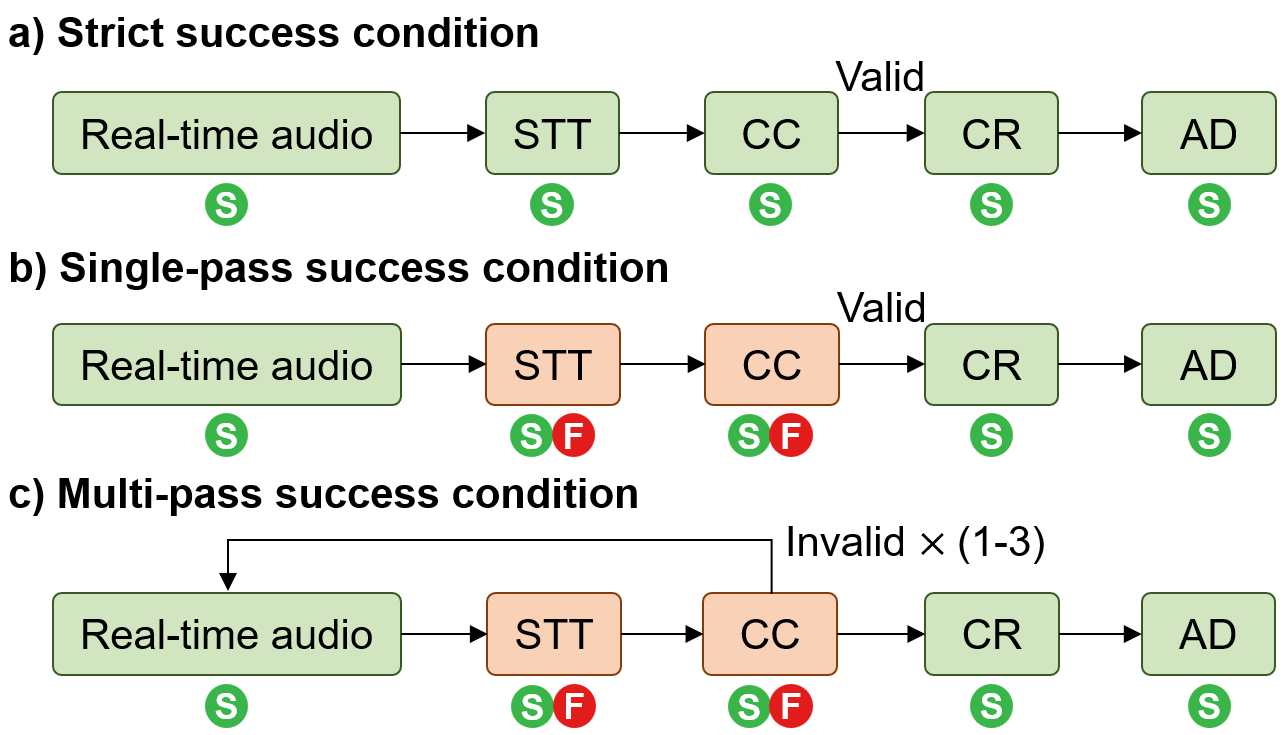}
    \caption{Success conditions for workflow-level SR, where green boxes represent success and red boxes denote success or fail: (a) In strict condition, the command is successful only if all stages succeed with no invalid loops; (b) In single-pass condition, it is also successful when intermediate failures are resolved in later stages, leading to a correct final outcome in a single attempt with no invalid loop; (c) In multi-pass condition, it is also successful when the final outcome is correct after multiple attempts, allowing up to three invalid loops.}
    \label{success_condition}
\end{figure}

The command-level orchestration evaluation measures two aspects: the stage-level accuracy that measures correctness at each workflow stage, and the workflow-level success rate (SR) that quantifies overall success throughout the entire workflow.
The stage-level accuracy is the mean of binary outcomes across all commands for each stage:
\begin{align}
\text{Accuracy}_{stage} = \frac{1}{N} \sum_{i=1}^{N} o_{i,stage}, \quad
o_{i,stage} \in \{0,1\} 
\label{Acc_stage}
\end{align} 
where $N$ denotes the total number of commands, 
$i$ indicates the command index, 
and $o_{i,stage}$ is the binary result of the $i$-th command at a stage.
The workflow-level SR evaluates each command under three conditions: strict, single-pass, and multi-pass as shown in Fig.~\ref{success_condition}. 
For each command $i$, a binary label $o_{i,condition} \in \{0,1\}$ indicates whether it satisfies the given success condition, and the $SR_{condition}$ is the average of these labels over all $N$ commands:
\begin{equation}
\text{SR}_{\text{condition}} = \frac{1}{N} \sum_{i=1}^{N} o_{i,\text{condition}}.
\label{SR_condition}
\end{equation}

The command-category evaluation examines how structural and linguistic variations affect orchestration performance and identifies which categories are more prone to errors.
For each category $c$, $SR_{multi-pass}$ is the mean of binary outcomes:
\begin{equation}
\text{SR}_{multi-pass}(c) = \frac{1}{N_c}\sum_{i \in c} o_{i,\text{multi-pass}},
\label{SR_cat}
\end{equation}
where $N_c$ is the number of commands in category $c$.
We also calculate the cross-category $SR_{multi-pass}$ for structure-type and type-expression pairs to explore how it jointly affects the performance.
For each pair of categories $(c_1, c_2)$, the cross-category $SR_{multi-pass}$ is:
\begin{equation}
\text{SR}_{\text{multi-pass}}(c_1, c_2) = \frac{1}{N_{c_1,c_2}} \sum_{i \in (c_1,c_2)} o_{i,\text{multi-pass}},
\label{SR_cross}
\end{equation}
where $N_{c_1,c_2}$ is the number of commands in the pair.

\begin{table*}[t]
\centering
\small
\caption{Example of VISA workflow stage outcomes for each command.}
\label{command_results}
\resizebox{\textwidth}{!}{
\begin{tabular}{clccccccccccc}
\hline
Agent & Command & Structure & Type & Expression & STT & CC & CR & AF & AP & AD & OF & IC \\
\hline
IR & Show patient information & Single & Explicit & Baseline & 1 & 1 & 1 & 1 & 1 & 1 & 1 & 0 \\
IR & Display the PFT info & Single & Explicit & Abbreviation & 1 & 1 & 1 & 1 & 1 & 1 & 1 & 0 \\
IR & Can you show pulmonary function test? & Single & NLQ & Baseline & 1 & 1 & 1 & 1 & 1 & 1 & 1 & 0 \\
IR & PFT results & Single & Implicit & Abbreviation & 1 & 1 & 1 & 1 & 1 & 1 & 1 & 0 \\
IR & lung function results & Single & Implicit & Paraphrase & 0 & 1 & 1 & 1 & 1 & 1 & 1 & 0 \\
IR & Show physical and PFT info & Composite & Explicit & Abbreviation & 1 & 1 & 1 & 1 & 1 & 1 & 1 & 0 \\
IR & Reset & Single & Implicit & Baseline & 1 & 1 & 1 & 1 & 1 & 1 & 1 & 0 \\
\hline
IV & Show the CT views & Single & Explicit & Baseline & 0 & 1 & 1 & 1 & 1 & 1 & 1 & 0 \\
IV & Coronal plus 100 & Single & Explicit & Baseline & 0 & 1 & 1 & 1 & 1 & 1 & 1 & 0 \\
IV & Sagittal minus 30 & Single & Explicit & Baseline & 1 & 1 & 1 & 1 & 1 & 1 & 1 & 0 \\
IV & Can you move CT forward? & Single & NLQ & Baseline & 1 & 1 & 1 & 1 & 1 & 1 & 1 & 0 \\
IV & Move front, front, front & Single & Implicit & Paraphrase & 0 & 1 & 1 & 1 & 1 & 1 & 1 & 0 \\
IV & Step to posterior & Single & Implicit & Paraphrase & 0 & 1 & 1 & 1 & 1 & 1 & 1 & 0 \\
IV & Axial zoom in & Single & Explicit & Baseline & 1 & 1 & 1 & 1 & 1 & 1 & 1 & 0 \\
IV & Can you get axial closer? & Single & NLQ & Paraphrase & 1 & 1 & 1 & 1 & 1 & 1 & 1 & 0 \\
IV & Minimize the axial image & Single & Explicit & Paraphrase & 1 & 1 & 1 & 1 & 1 & 1 & 1 & 0 \\
IV & Could you reduce coronal image? & Single & NLQ & Paraphrase & 0 & 1 & 1 & 0 & 0 & 0 & 1 & 0 \\
IV & Zoom out & Single & Implicit & Baseline & 1 & 1 & 1 & 1 & 1 & 1 & 1 & 0 \\
IV & Move axial to the middle slice and zoom in & Composite & Explicit & Paraphrase & 1 & 1 & 1 & 1 & 1 & 1 & 1 & 0 \\
IV & Can you move axial to 200 and coronal to 230? & Composite & NLQ & Baseline & 0 & 1 & 1 & 1 & 1 & 1 & 1 & 0 \\
\hline
AR & Show the 3D recon image & Single & Explicit & Baseline & 1 & 1 & 1 & 1 & 1 & 1 & 1 & 0 \\
AR & Can you load anatomical reconstruction? & Single & NLQ & Paraphrase & 1 & 1 & 1 & 1 & 1 & 1 & 1 & 1 \\
AR & Turn on RLL & Single & Explicit & Abbreviation & 1 & 1 & 1 & 1 & 1 & 1 & 1 & 0 \\
AR & Can you hide left lung? & Single & NLQ & Paraphrase & 1 & 1 & 1 & 1 & 1 & 1 & 1 & 0 \\
AR & Show anterior view & Single & Explicit & Baseline & 1 & 1 & 1 & 1 & 1 & 1 & 1 & 0 \\
AR & Can you look from the front? & Single & NLQ & Paraphrase & 1 & 1 & 1 & 1 & 1 & 1 & 1 & 0 \\
AR & Surgical view & Single & Implicit & Baseline & 1 & 1 & 1 & 1 & 1 & 1 & 1 & 0 \\
AR & Surgeon's view & Single & Implicit & Paraphrase & 1 & 1 & 1 & 1 & 1 & 1 & 1 & 0 \\
AR & Rotate to the right & Single & Explicit & Baseline & 1 & 1 & 1 & 1 & 1 & 1 & 1 & 0 \\
AR & Would you rotate up? & Single & NLQ & Baseline & 1 & 1 & 1 & 1 & 1 & 1 & 1 & 0 \\
AR & Can you zoom in to RLL? & Single & NLQ & Abbreviation & 1 & 1 & 1 & 0 & 1 & 0 & 1 & 0 \\
AR & Zoom out & Single & Implicit & Baseline & 1 & 1 & 1 & 1 & 0 & 0 & 1 & 0 \\
AR & Can you zoom in and rotate to the left? & Composite & NLQ & Baseline & 1 & 1 & 1 & 1 & 1 & 1 & 1 & 0 \\
AR & Can you initialize and zoom in? & Composite & NLQ & Baseline & 1 & 1 & 1 & 1 & 1 & 1 & 1 & 0 \\
AR & Can you erase all? & Single & NLQ & Paraphrase & 1 & 1 & 1 & 1 & 1 & 1 & 1 & 0 \\
\hline
\multicolumn{13}{p{1.1\textwidth}}{All results available at \url{https://github.com/helena-lena/SAOP}.}
\end{tabular}
}
\end{table*}

\section{Results}
\subsection{Command-level Orchestration Evaluation}

\subsubsection{Stage-level Accuracy}

\begin{figure*}[t]
    \centering
    \includegraphics[width=0.8\textwidth]{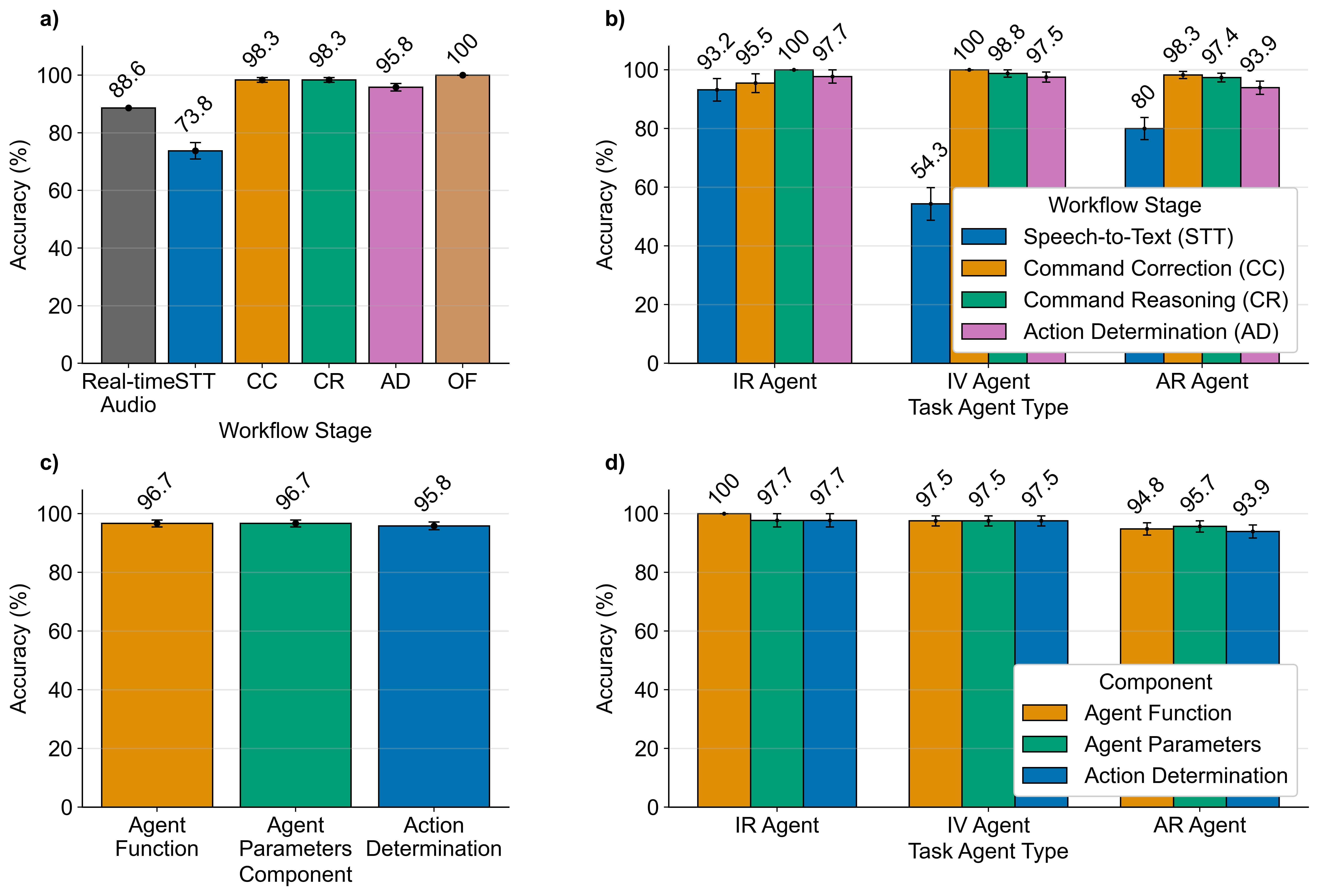}
    \caption{Stage-level accuracy results: (a) Overall stage-level accuracy; (b) Stage-level accuracy by task agent type; (c) Overall accuracy of action determination components; (d) Action determination component accuracy by task agent type; Error bars indicate 95\% confidence intervals.}
    \label{stage_level_accuracy}
\end{figure*} 

\begin{figure*}[t]
    \centering
    \includegraphics[width=0.75\linewidth]{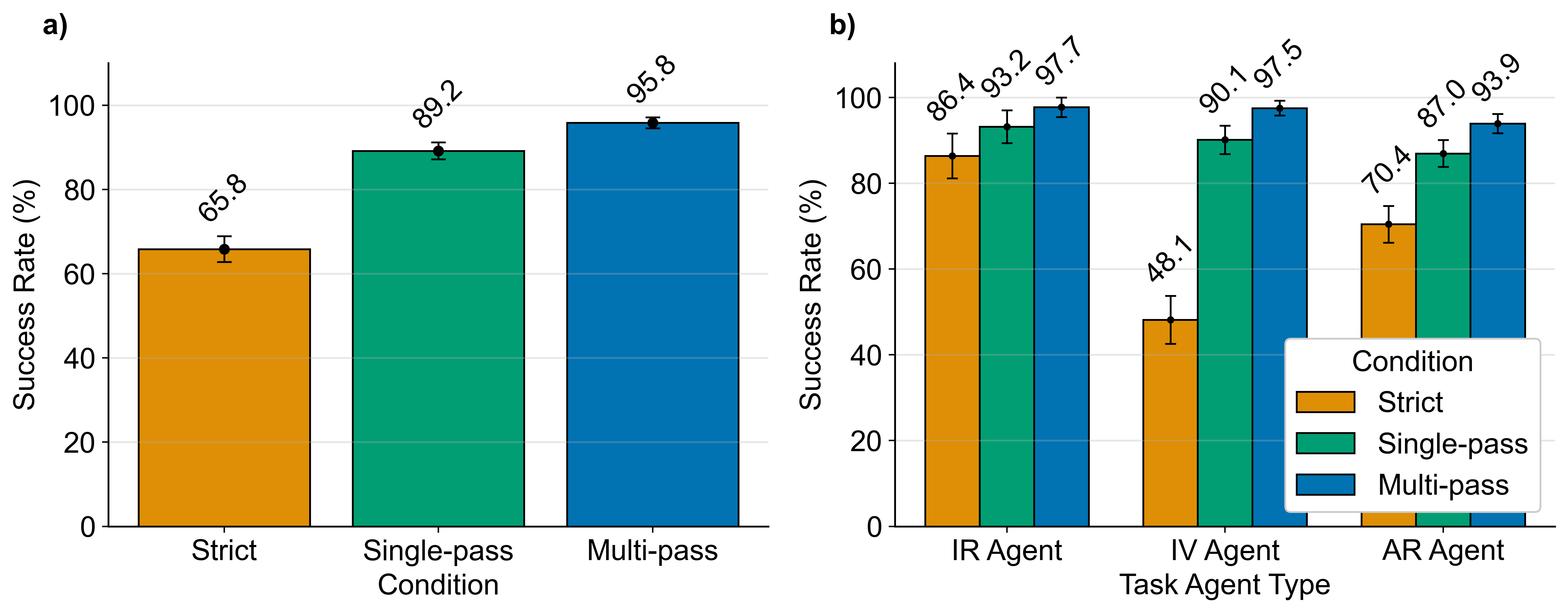}
    \caption{Workflow-level sucess rates results: (a) Overall results; (b) Results by task agent type; Error bars indicate 95\% confidence intervals.}
    \label{workflow_level_success_rate}
\end{figure*} 

The VISA demonstrates strong error recovery capability from STT stage failures, maintaining high accuracy across subsequent stages as shown in Fig.~\ref{stage_level_accuracy} (a).
The real-time audio stage accuracy is the performance of the wake-word detection model for the keyword “davinci”, which correctly recognized 266 out of 300 attempts ($88.6\%$).
The remaining 34 attempts correspond to a false rejection rate of $11.3\%$, mainly due to the model misinterpreting “davinci” such as “Dabinci”, “David”, “Benchi”, or “dubbing”.
For successful detections, the latency between speech detection and the “davinci” recognition averaged 1.97 seconds, showing smooth real-time interaction.
Although STT accuracy is lower due to transcription errors, most are corrected in the CC stage.
The CR stage maintains high accuracy in selecting the agent, AD accuracy shows a slight decline due to occasional mispredictions in the agent function or parameters.
Finally, the OF achieves $100\%$ accuracy, confirming that all workflow stages execute in the correct order.

The stage-level accuracies across the three task-specific agents exhibit similar trends (Fig.~\ref{stage_level_accuracy}b).
The STT stage shows the largest variation, with the IV agent exhibiting the lowest accuracy due to frequent misrecognition of “CT” as “city”.
Nevertheless, most recognition errors are corrected during the CC stage, the subsequent CR stage also maintains high performance across all agents, and the AD stage shows a slight decrease.
Later stages generally show lower accuracy than the CC stage, reflecting the natural accumulation of prediction errors when performing stages step-by-step with an LLM.
Interestingly, the IR agent even compensates for imperfectly corrected commands during the CR stage.
For example, when the STT transcribes “age info” as “Age in full”, the CC stage refines it to “Age information in full”, allowing the CR stage to infer the intent as a request for the age column only.
Overall, VISA demonstrates strong robustness, effectively mitigating early-stage errors and ensuring reliable agent execution.

In the AD stage, each task-specific agent determines the action uniquely mapped to an agent function, and its parameters specifying how the action should be executed.
Our VISA achieves consistently high accuracy for the function and parameters, and because the mispredictions in the two components do not overlap, the combined AD accuracy is slightly lower (Fig.~\ref{stage_level_accuracy}c).
The accuracies across the three task agents exhibit similar trends, where the integrated AD accuracy is comparable to or slightly lower than the accuracies of function and parameter predictions (Fig.~\ref{stage_level_accuracy}d).
A slight decrease in accuracy from the IR to IV and AR agents is likely due to the greater variability and complexity of the commands in the latter tasks.
Nevertheless, the differences remain marginal, indicating that the LLM effectively generalizes its reasoning capabilities across heterogeneous task agents.

\subsubsection{Workflow-level Success Rate}

The workflow-level SR improves progressively from strict to single-pass to multi-pass conditions as shown in Fig.~\ref{workflow_level_success_rate} (a).
The $SR_{strict}$ remains at 65.8\%, primarily limited by transcription errors in the STT.
In contrast, the $SR_{single-pass}$ increases to 89.2\% as the LLM resolves early-stage failures through correction and reasoning.
The $SR_{multi-pass}$ reaches 95.8\%, showing that the workflow orchestrator agent returns to the real-time audio stage for invalid commands, allowing the user to restate the command more clearly.
A similar trend appears in task-specific agents as shown in Fig.~\ref{workflow_level_success_rate} (b).
The IV agent shows the largest gap between strict and single-pass due to frequent STT errors, whereas the IR shows the smallest.
As task complexity increases from IR to IV and AR, both $SR_{single-pass}$ and $SR_{multi-pass}$ decrease.

\begin{figure*}[t]
    \centering
    \includegraphics[width=\textwidth]{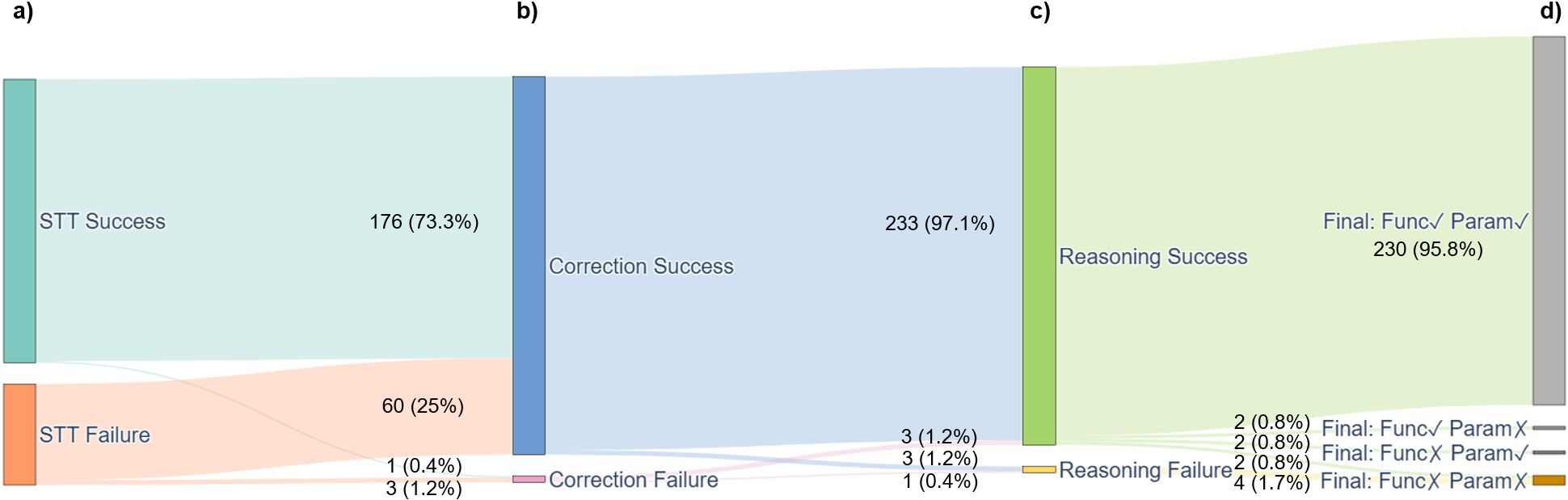}
    \caption{A sankey diagram illustrating recovery and failure paths of the command through stages from (a) STT, (b) CC, (c) CR, and (d) AD.}
    \label{failure_recovery_flow_details}
\end{figure*} 

\subsubsection{Failure and Recovery Paths}

We also analyze command processing paths from STT to AD stages as shown in Fig.~\ref{failure_recovery_flow_details}.
Most STT errors arise from confusing medical terms with phonetically similar non-medical words, such as recognizing “CT” as “city”, “coronal” as “corona”, and “lung” as “long”.
Minor lexical errors also occur like recognizing “to” as “2”, “right” as “write”, “zoom” as “June”, and “add” as “at”.
While most are resolved in the CC stage, a few remain, converting medical terms into non-medical words or nouns into adjective forms.
CR failures occur when commands remain invalid after correction, whereas minor CC errors often do not affect agent selection, resulting in recovery.
AD failures mainly arise from composite commands or mispredictions on functions or parameters.
These findings highlight the need to support multi-step actions and refine prompts for challenging cases.

\subsection{Command-category Evaluation}

\begin{figure*}[t]
    \centering
    \includegraphics[width=0.8\textwidth]{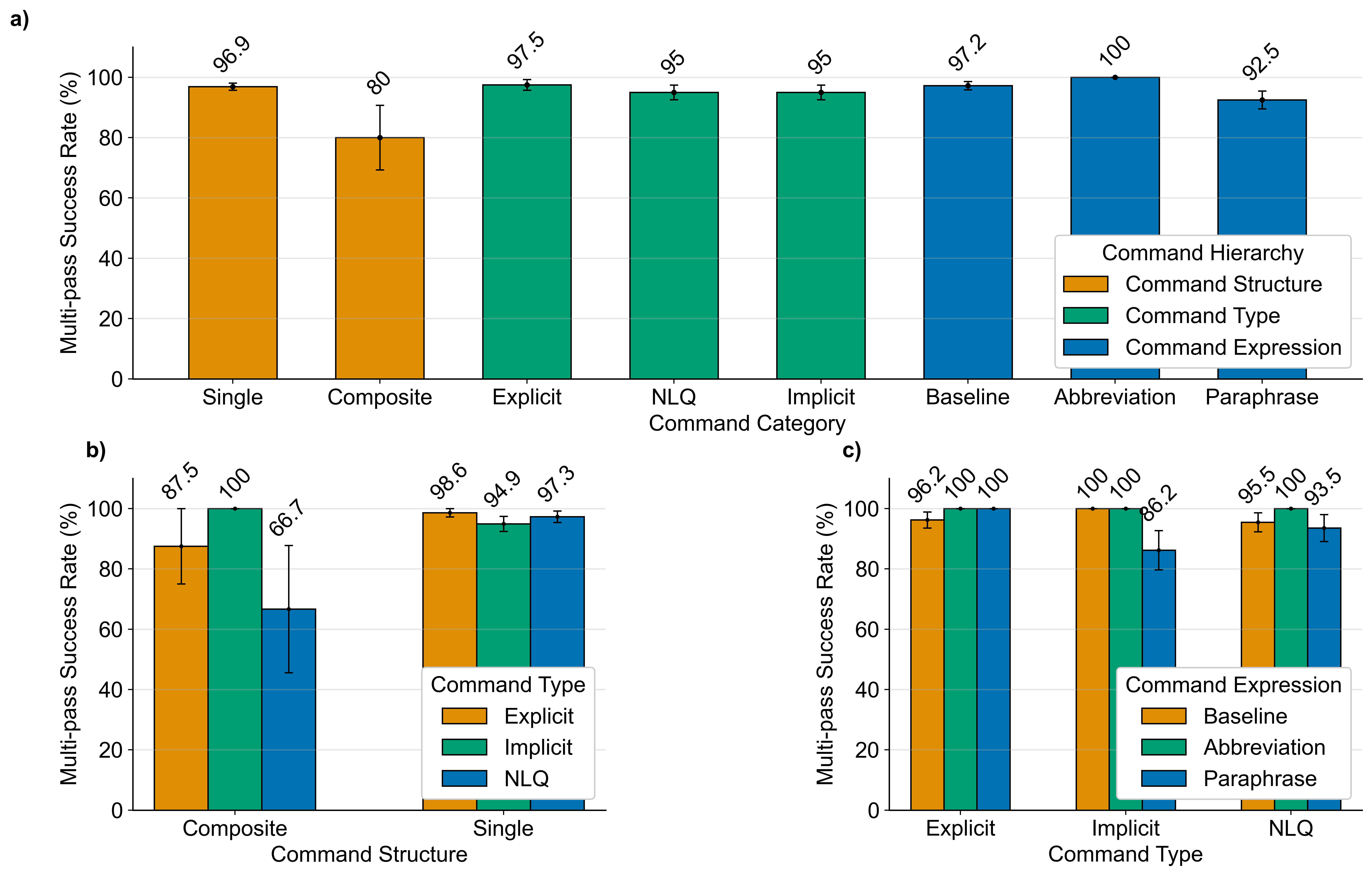}
    \caption{Multi-pass success rate across command categories: (a) Multi-pass success rate by command category; (b) Multi-pass success rate by structure $\times$ type; (c) Multi-pass success rate by type $\times$ expression; Error bars in all plots indicate 95\% confidence intervals.}
    \label{multi_pass_success_rate_by_categories}
\end{figure*}

We evaluate $SR_{multi-pass}$ across hierarchical command categories and VISA robustly handles a wide range of structural and linguistic variations in user commands, as shown in Fig.~\ref{multi_pass_success_rate_by_categories} (a).
Within the command structure category, composite commands are more challenging than single commands because the platform struggles to determine the correct agent function when multiple functions must be executed sequentially.
For the command type category, the platform shows strong performance for explicit commands, implicit commands, and NLQ, demonstrating that the LLM can follow user intent with minimal performance degradation despite linguistic variations.
Regarding the command expression category, commands using abbreviations achieve perfect performance because the LLM prompt explicitly includes the medical domain abbreviations.
Commands with paraphrased expressions show slightly lower performance, as such variations are not included in the prompt and may introduce additional ambiguity.

\begin{figure*}[t]
    \centering
    \includegraphics[width=\textwidth]{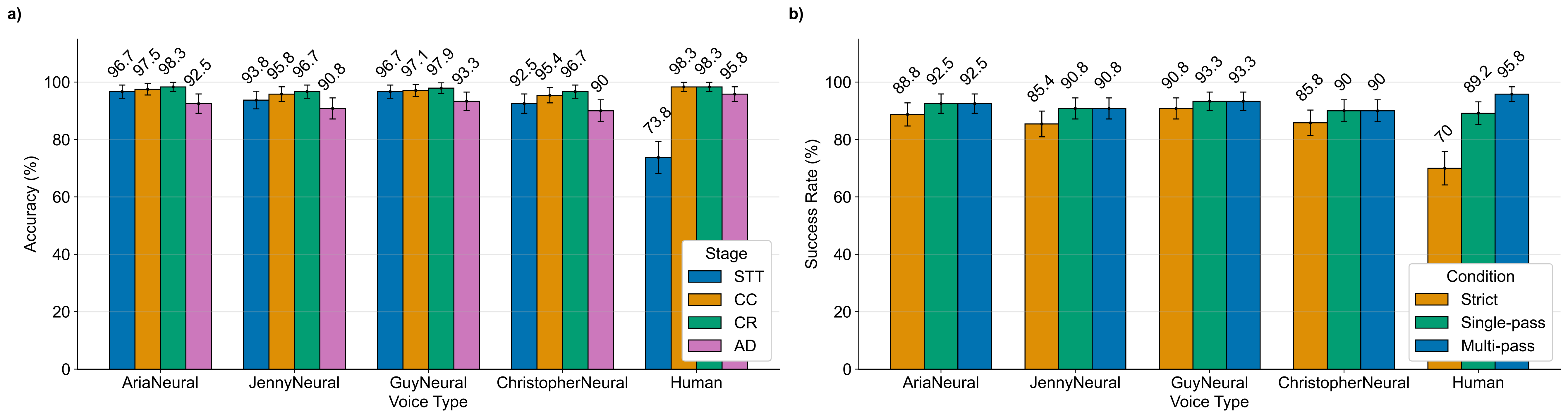}
    \caption{Comparison by voice type: (a) Stage-level accuracy by voice type; (b) Success rate by voice type; Error bars are 95\% confidence intervals.}
    \label{comparison_by_voice_type}
\end{figure*} 

We further analyze cross-category $SR_{multi-pass}$ as illustrated in Fig.~\ref{multi_pass_success_rate_by_categories} (b) and (c).
In the structure–type relationship, single commands generally outperform composite ones, with composite failures occurring in three cases that require multiple actions, such as “Can you zoom in and rotate to the left?” or “Can you initialize and zoom in?”.
All command types under the single structure maintain high performance, demonstrating that the VISA reliably processes straightforward, single-step commands.
In the type–expression relationship, explicit commands achieve strong performance across all expression forms, whereas and only two failures arise from composite structure and entering invalid loops.
In contrast, implicit commands are less robust to unseen synonyms or paraphrases, yet still maintain reasonable performance.
Three failures are due to entering invalid loops, and one results from misinterpreting “pre-existing condition” and displaying both “comorbidities” and “diagnosis”.
NLQs perform slightly better with paraphrase expressions because they usually contain richer contextual detail than short and implicit commands.
Two failures appear: one when the VISA interprets “Could you reduce coronal image?” as adjusting slice position instead of zooming out, and another when it interprets “Can you view the superior angle?” as rotating upward rather than switching to the superior view.
These results show that VISA generalizes well across different categories, with performance decreasing for composite or paraphrased commands that require multi-action and deeper reasoning on unseen expressions.

We further investigate the successful cases among challenging categories, composite commands and commands in paraphrase format.
For composite commands, the IR agent retrieves all requested data in cases like “show physical and PFT info” and “show surgery and tumor information”.
The IV agent accurately executes commands like “move axial to the middle slice and zoom in” or “sagittal to the center slice and zoom in”, computing mid-slice positions automatically with zoom in action, and correctly handles multi-plane commands such as “move to axial 230, coronal 220, sagittal 220” or “move to 10, 20, 50” even without explicit plane order.
The AR agent correctly executes composite commands like “view from the front and rotate left” or “view from the posterior and rotate right”, when a single function can handle both actions through parameter adjustment.
For paraphrased commands, we examine cases where such expressions are not explicitly defined in the prompt but convey semantically similar meanings.
The IR agent can handle expressions such as retrieving the age for “How old is the patient?” and the sex for “What’s the gender?”.
The IV agent interprets various expressions, executing commands like “Move front front front”, “Move left left left left”, and “Move down down” as movements to “Coronal plus 30”, “Sagittal minus 40”, and “Axial minus 20”.
As the prompt defines the default movement distance for 10 units, the LLM infers the repetitions to multiples of that distance. 
It also predicts positions correctly for “Coronal to maximum”, “minimum”, and “middle slice”, and adapts to unseen verbs such as “step” or “slide”.
For zoom operations, it performs “Zoom in” and “Zoom out” correctly for paraphrases like “Coronal enlarge,” “Focus on coronal plane,” or “Return to smaller view.”
The AR agent recognizes “Add airway” as the trachea and bronchia, sets the anterior view for “look from the front,” and correctly interprets unseen synonyms for rotation and zoom actions.
Overall, these results show that the LLM handles synonymous and paraphrased expressions with strong linguistic flexibility.

\subsection{Applicability for Real Surgical Settings}

In surgical environments, voice characteristics differ across speakers in tone, accent, speech rate, and pronunciation.
To evaluate robustness to speaker variability, we compare four synthesized neural voices with human speech as shown in Fig.~\ref{comparison_by_voice_type}.
Synthesized voices achieve higher STT accuracy due to their clear and native-like pronunciation, whereas human speakers recover more effectively from invalid loops by restating the command more clearly and emphasizing key information.
Although some reasoning errors occur during the AD stage, all voice types maintain high stage-level accuracy.
The workflow-level SR under strict, single-pass, and multi-pass conditions exhibit similar trends.
Human speech exhibits lower $SR_{strict}$ because of STT failures, but ultimately achieves the highest $SR_{multi-pass}$ by recovering from invalid commands. 
These results demonstrate that VISA maintains stable performance across diverse voice conditions.

Furthermore, the operating room contains background noise and conversations from multiple surgeons and nurses.
To ensure that wake-word “davinci” detection does not activate accidentally, we evaluate $15$ unsupported command examples as shown in Table~\ref{non_command_examples}.
The wake-word detection module remains active for one hour, and an unsupported command is spoken every four minutes.
During one hour of testing, the VISA records a false alarm rate of 0, indicating no unintended activations.
We also evaluate whether these $15$ unsupported commands are correctly classified as invalid after the wake word “davinci” is triggered.
Three are misclassified as valid patient information requests, asking about vital sign, $CO_{2}$ pressure, and frozen section.
Since these fields are not present in the given data columns, the VISA either overlays the previous video information or hides the overlay.
Overall, our VISA reliably distinguishes task-related commands from unsupported commands, though a few cases still require refinement to prevent incorrect validation.

\begin{table}[t]
\centering
\caption{Results on unsupported commands}
\resizebox{\columnwidth}{!}{
\begin{tabular}{@{}lcc@{}}
\hline
Example & Wake-word & Invalid \\
\hline
Replace the gauze. & No & Yes \\
Apply suction. & No & Yes \\
Clean the camera lens. & No & Yes \\
Wipe the third arm. & No & Yes \\
Switch between arm 1 and 2. & No & Yes \\
Prepare the hemostatic agent. & No & Yes \\
How long has the surgery been going on? & No & Yes \\
Is the patient’s vital sign stable? & No & No \\
What is the current CO\textsubscript{2} pressure? & No & No \\
How long since the frozen section was sent? & No & No \\
The patient is coughing. & No & Yes \\
Prepare the stapler. & No & Yes \\
Prepare the ICG injection. & No & Yes \\
Prepare for the air leak test. & No & Yes \\
Insert the retrieval bag. & No & Yes \\
\hline
\end{tabular}
}
\label{non_command_examples}
\end{table}

\subsection{Analysis of Sequential Command Execution}

\begin{figure*}[t]
    \centering
    \includegraphics[width=\textwidth]{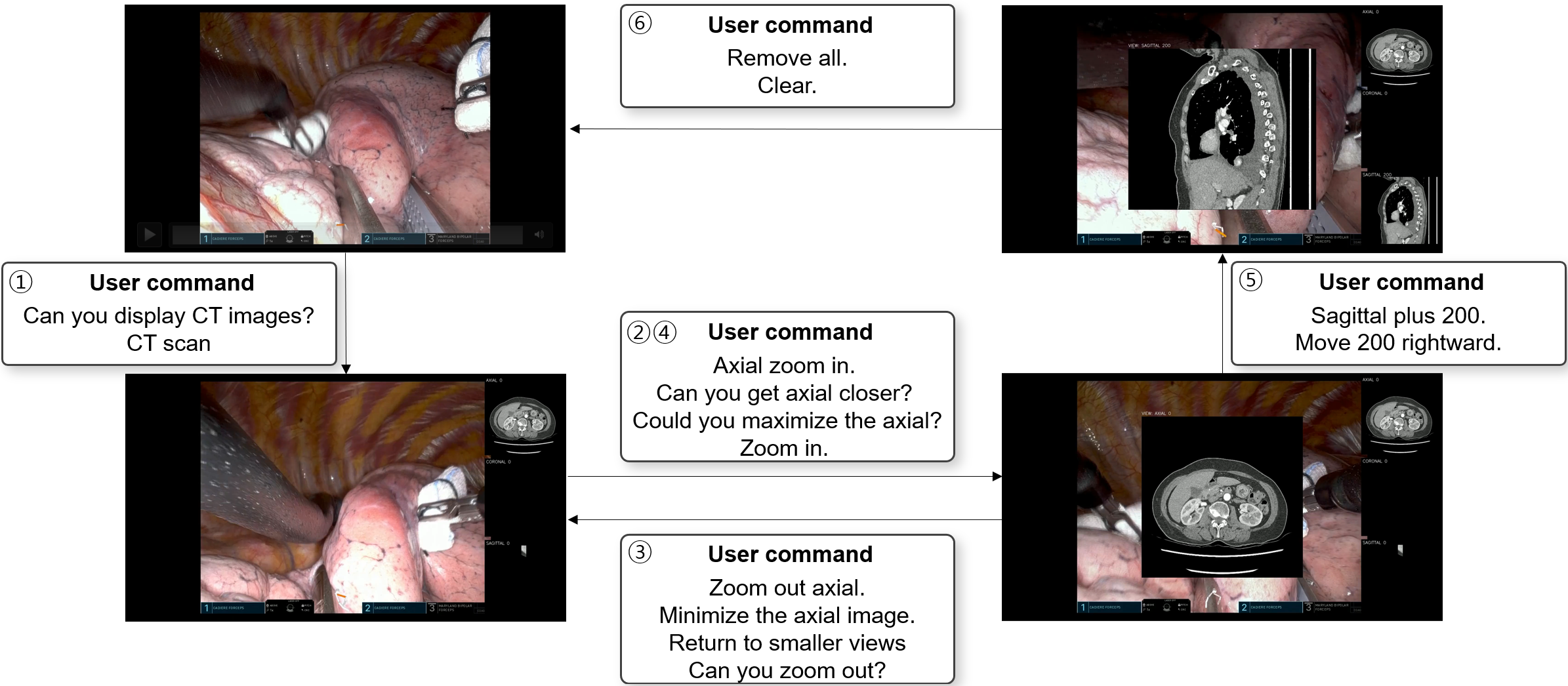}
    \caption{Video examples of consecutive scenarios for the IV agent.}
    \label{CT_agent_examples}
\end{figure*}

\begin{figure*}[t]
    \centering
    \includegraphics[width=0.9\textwidth]{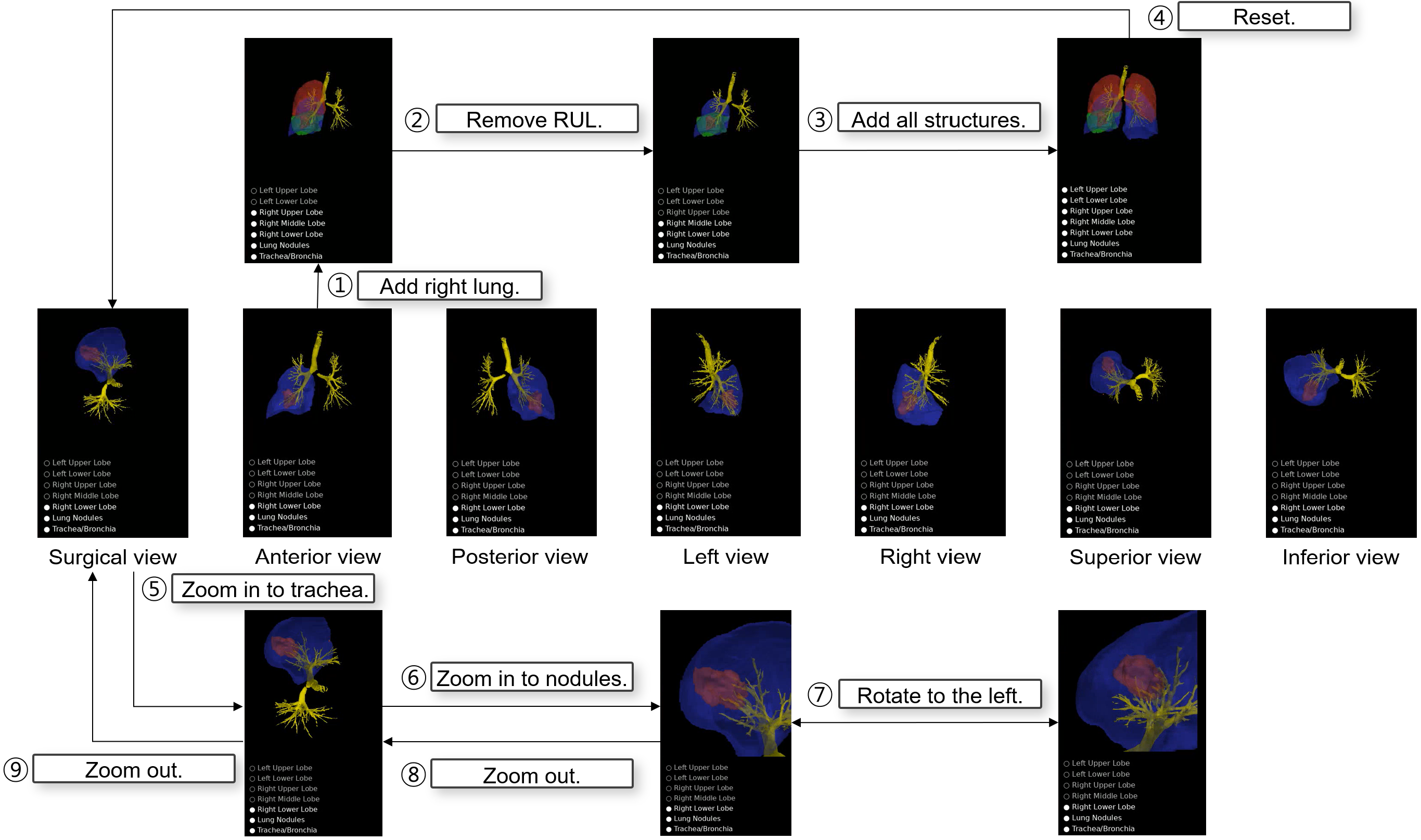}
    \caption{Video examples of consecutive scenarios for the AR agent.}
    \label{Recon_agent_examples}
\end{figure*}

We further evaluate whether VISA maintains continuity across sequential video clips\footnote{Videos available at \url{https://helena-lena.github.io/SAOP/}} using its memory state, as shown in Fig.~\ref{CT_agent_examples} and Fig.~\ref{Recon_agent_examples}.
The IV agent continuously updates axial, coronal, and sagittal slice positions within the predefined minimum and maximum constraints.
For the AR agent, VISA preserves continuity in zoom operations by accurately computing the starting and ending scales while maintaining the minimum scale constraint of 1.
The rotation and view adjustments also function correctly within the zoomed state, and the zoom out action follows the exact reverse path of zoom in.
Additionally, when no command appears in the target video clip, the CC stage revises the empty input to “Select {agent}”, and the AD stage restores the previous agent state.
For the “Reset” command, each agent restores the default settings of its function and the parameters defined in the prompt.
These results demonstrate that VISA preserves temporal and functional consistency across video clips, ensuring stable and continuous state transitions.

\section{Discussion} 

We demonstrate the effectiveness of the hierarchical multi-agent framework with LLM-based orchestration.
The workflow orchestrator agent reliably plans and selects appropriate functions based on the current state and predefined decision rules.
The hybrid design of LLM-driven reasoning with probability-based decision rules resolves invalid loops and enables workflow stages to progress smoothly.
Although computing probabilities for all functions slows inference compared to direct function prediction, it could enable conditional probability–based planning to handle increasingly complex paths as the number of agents and functions grows.
Separating workflow functions from task-specific agent functions further improves modularity and scalability, allowing flexible integration of new agents or functions.
Expanding clinically useful agents, such as detecting critical blood vessels, supporting intraoperative navigation, and assisting in robotic planning, will enhance the clinical applicability.

While our VISA demonstrates the feasibility of LLM-based orchestration for surgical assistance, the observed errors reveal opportunities for improvement.
At the STT stage, most errors occur with medical domain-specific terminology, suggesting the need to fine-tune lightweight STT models on medical terms to improve transcription accuracy.
In the CC stage, the LLM revises misrecognized words to their intended medical terms, yet augmenting text correction rules or fine-tuning a correction model on medical context could further enhance robustness in failure cases.
At the AD stage, most errors arise from composite commands, indicating the need to extend the VISA to support multi-step sequential operations.
Additionally, since the current prompting strategy are closely coupled with the Gemma3~\cite{Gemma3:2025} model, performance may vary across different LLMs.
Developing a self-evolving orchestration mechanism that learns from prior errors and dynamically refines its prompts could ensure adaptability and consistent performance across models in future implementations.

From a dataset perspective, the evaluation uses 240 commands, including both human and synthesized voice inputs from four speakers to simulate variability.
However, the dataset does not fully represent real clinical variability in phrasing, terminology, or multilingual mixing observed in real clinical settings.
Future work may involve fine-tuning models on multilingual medical data or using fast and commercially available multilingual STT systems.
Moreover, the clinical information displayed by the IR agent is based on a predefined set of columns selected by clinicians.
However, the required fields may vary across departments and individual surgeons, indicating the need for customizable data configurations.
For 3D anatomical models displayed by AR agent, using commercial or clinically validated software could provide more precise anatomical structures and support more sophisticated manipulation for clinical use.

Regarding computational efficiency, we employ Gemma3 27B (\text{gemma3:27b-it-qat}) as the core LLM for all VISA stages, a quantization-aware trained version that preserves high performance with reduced memory usage.
We set the temperature to 0 to minimize randomness, use \text{num\_predict} of $-1$ for unrestricted generation, and fix the seed at 42 for reproducibility.
The model runs locally via Ollama\footnote{\url{https://github.com/ollama/ollama}} and occupies approximately 17 GB when preloaded onto a single NVIDIA RTX 4500 Ada GPU (24 GB), while the second GPU handles other APIs, models, or agents.
However, cumulative latency from sequential LLM calls remains a challenge in time-sensitive surgical environments.
This delay becomes more pronounced when the LLM engages in complex reasoning or generates longer output.
Latency may be reduced by limiting LLM reasoning to essential stages, caching reasoning patterns, shortening outputs, and adapting to surgeon-specific interaction patterns.
Lastly, for 3D model rendering, increasing mesh decimation or frame skipping improves speed but may reduce smoothness during rotation and zoom.

\appendices


\section*{References}

\begin{thebibliography}{00}
\bibitem{allaf1998laparoscopic} 
M. E. Allaf, S. V. Jackman, P. G. Schulam, J. A. Cadeddu, B. R. Lee, R. G. Moore {\it et al.}, ``Laparoscopic visual field: Voice vs foot pedal interfaces for control of the {AESOP} robot,'' {\it Surg. Endosc.}, vol. 12, no. 12, pp. 1415--1418, 1998.

\bibitem{mettler1998one}
L. Mettler, M. Ibrahim, and W. Jonat, ``One year of experience working with the aid of a robotic assistant (the voice-controlled optic holder {AESOP}) in gynaecological endoscopic surgery,'' {\it Hum. Reprod.}, vol. 13, no. 10, pp. 2748--2750, 1998.

\bibitem{reichenspurner1999use}
H. Reichenspurner, R. J. Damiano, M. Mack, D. H. Boehm, H. Gulbins, C. Detter {\it et al.}, ``Use of the voice-controlled and computer-assisted surgical system {ZEUS} for endoscopic coronary artery bypass grafting,'' {\it J. Thorac. Cardiovasc. Surg.}, vol. 118, no. 1, pp. 11--16, 1999.

\bibitem{nathan2006voice}
C. O. Nathan, V. Chakradeo, K. Malhotra, H. D'Agostino, and R. Patwardhan, ``The voice-controlled robotic assist scope holder {AESOP} for the endoscopic approach to the sella,'' {\it Skull Base}, vol. 16, no. 3, pp. 123--131, 2006.

\bibitem{salama2005utility}
I. A. Salama and S. D. Schwaitzberg, ``Utility of a voice-activated system in minimally invasive surgery,'' {\it J. Laparoendosc. Adv. Surg. Tech.}, vol. 15, no. 5, pp. 443--446, 2005.

\bibitem{vaida2010development}
C. Vaida, D. Pisla, N. Plitea, B. Gherman, B. Gyurka, F. Graur {\it et al.}, ``Development of a voice controlled surgical robot,'' in {\it New Trends in Mechanism Science: Analysis and Design}. Dordrecht, The Netherlands: Springer, 2010, pp. 567--574. [Online]. Available: https://link.springer.com/chapter/10.1007/978-90-481-9689-0\_65?utm\_source=chatgpt.com

\bibitem{forte2022design}
M. P. Forte, R. Gourishetti, B. Javot, T. Engler, E. D. Gomez, and K. J. Kuchenbecker, ``Design of interactive augmented reality functions for robotic surgery and evaluation in dry-lab lymphadenectomy,'' {\it Int. J. Med. Robot.}, vol. 18, no. 2, p. e2351, 2022.

\bibitem{elazzazi2022natural}
M. Elazzazi, L. Jawad, M. Hilfi, and A. Pandya, ``A natural language interface for an autonomous camera control system on the da Vinci surgical robot,'' {\it Robotics}, vol. 11, no. 2, p. 40, 2022.

\bibitem{kim2025speech}
Y. G. Kim, J. W. Shim, G. Gimm, S. Kang, W. Rhee, J. H. Lee {\it et al.}, ``Speech-mediated manipulation of da Vinci surgical system for continuous surgical flow,'' {\it Biomed. Eng. Lett.}, vol. 15, no. 1, pp. 117--129, 2025.

\bibitem{davila2024voice}
A. Davila, J. Colan, and Y. Hasegawa, ``Voice control interface for surgical robot assistants,'' in {\it Proc. 2024 Int. Symp. Micro-NanoMehatronics Hum. Sci.} IEEE, 2024, pp. 1--5.

\bibitem{pandya2023chatgpt}
A. Pandya, ``{ChatGPT}-enabled daVinci surgical robot prototype: Advancements and limitations,'' {\it Robotics}, vol. 12, no. 4, p. 97, 2023.

\bibitem{zhang2025llm}
Y. Zhang, B. Orthmann, M. C. Welle, J. Van Haastregt, and D. Kragic, ``{LLM}-driven augmented reality puppeteer: Controller-free voice-commanded robot teleoperation,'' in {\it Proc. Int. Conf. Human-Computer Interaction}. Springer, 2025, pp. 97--112.

\bibitem{gavric2025surgery}
A. Gavric, D. Bork, and H. A. Proper, ``Surgery {AI}: Multimodal process mining and mixed reality for real-time surgical conformance checking and guidance,'' {\it ZEUS 2025}, p. 48, 2025.

\bibitem{Touvron2023}
OpenAI, ``{GPT-4} technical report,'' arXiv:2303.08774, 2023.

\bibitem{Chowdhery2023}
A. Chowdhery, S. Narang, J. Devlin, M. Bosma, G. Mishra, A. Roberts {\it et al.}, ``{PaLM}: Scaling language modeling with pathways,'' {\it J. Mach. Learn. Res.}, vol. 24, no. 240, pp. 1--113, 2023.

\bibitem{Touvron:2023:LLaMA}
H. Touvron, T. Lavril, G. Izacard, X. Martinet, M.-A. Lachaux, T. Lacroix {\it et al.}, ``{LLaMA}: Open and efficient foundation language models,'' arXiv:2302.13971, 2023.

\bibitem{Touvron:2023:LLaMA2}
H. Touvron, L. Martin, K. Stone, P. Albert, A. Almahairi, Y. Babaei {\it et al.}, ``{LLaMA} 2: Open foundation and fine-tuned chat models,'' arXiv:2307.09288, 2023.

\bibitem{Dubey:2024:LLaMA3}
A. Grattafiori, A. Dubey, A. Jauhri, A. Pandey, A. Kadian, A. Al-Dahle {\it et al.}, ``The {LLaMA} 3 herd of models,'' arXiv:2407.21783, 2024.

\bibitem{Meta:2025:LLaMA4}
Meta, ``The {LLaMA} 4 herd: The beginning of a new era of natively multimodal {AI} innovation,'' 2025. [Online]. Available: https://ai.meta.com/blog/llama-4-multimodal-intelligence/

\bibitem{Gemma:2024}
Gemma Team, ``{Gemma}: Open models based on gemini research and technology,'' arXiv:2403.08295, 2024.

\bibitem{Gemma2:2024}
Gemma Team, ``{Gemma} 2: Improving open language models at a practical size,'' arXiv:2408.00118, 2024.

\bibitem{Gemma3:2025}
Gemma Team, ``{Gemma} 3 technical report,'' arXiv:2503.19786, 2025.

\bibitem{Wei2022}
J. Wei, X. Wang, D. Schuurmans, M. Bosma, B. Ichter, F. Xia {\it et al.}, ``Chain-of-thought prompting elicits reasoning in large language models,'' in {\it Proc. Adv. Neural Inf. Process. Syst.}, vol. 35. NeurIPS, 2022, pp. 24\,824--24\,837.

\bibitem{wang2022scCoT}
X. Wang, J. Wei, D. Schuurmans, Q. Le, E. Chi, S. Narang {\it et al.}, ``Self-consistency improves chain of thought reasoning in language models,'' arXiv:2203.11171, 2022.

\bibitem{yao2023ToT}
S. Yao, D. Yu, J. Zhao, I. Shafran, T. Griffiths, Y. Cao {\it et al.}, ``Tree of thoughts: Deliberate problem solving with large language models,'' in {\it Proc. Adv. Neural Inf. Process. Syst.}, vol. 36. NeurIPS, 2023, pp. 11\,809--11\,822.

\bibitem{besta2024GoT}
M. Besta, N. Blach, A. Kubicek, R. Gerstenberger, M. Podstawski, L. Gianinazzi {\it et al.}, ``Graph of thoughts: Solving elaborate problems with large language models,'' in {\it Proc. AAAI Conf. Artif. Intell.}, vol. 38. AAAI, 2024, pp. 17\,682--17\,690.

\bibitem{franklin1996agent}
S. Franklin and A. Graesser, ``Is it an agent, or just a program?: A taxonomy for autonomous agents,'' in {\it Int. Workshop Agent Theories, Architectures, Lang.} Springer, 1996, pp. 21--35.

\bibitem{russell1995modern}
S. Russell and P. Norvig, ``Artificial intelligence,'' in {\it A Modern Approach}. Englewood Cliffs, NJ, USA: Prentice Hall, 1995, pp. 31--33.

\bibitem{huang2025introduction}
K. Huang and C. Hughes, ``Introduction to agentic {AI}: Foundations, drivers, and risks,'' in {\it Securing AI Agents: Foundations, Frameworks, and Real-World Deployment}. Cham, Switzerland: Springer, 2025, pp. 3--16.

\bibitem{yao2022react}
S. Yao, J. Zhao, D. Yu, N. Du, I. Shafran, K. R. Narasimhan {\it et al.}, ``{ReAct}: Synergizing reasoning and acting in language models,'' in {\it Proc. Int. Conf. Learn. Representations}. ICLR, 2022.

\bibitem{wu2024autogen}
Q. Wu, G. Bansal, J. Zhang, Y. Wu, B. Li, E. Zhu {\it et al.}, ``{AutoGen}: Enabling next-gen {LLM} applications via multi-agent conversations,'' in {\it Proc. First Conf. Lang. Modeling}, 2024.

\bibitem{li2023camel}
G. Li, H. Hammoud, H. Itani, D. Khizbullin, and B. Ghanem, ``{CAMEL}: Communicative agents for ``mind'' exploration of large language model society,'' in {\it Proc. Adv. Neural Inf. Process. Syst.}, vol. 36. NeurIPS, 2023, pp. 51\,991--52\,008.

\bibitem{zhang2025survey}
K. Zhang, Y. Zuo, B. He, Y. Sun, R. Liu, C. Jiang {\it et al.}, ``A survey of reinforcement learning for large reasoning models,'' arXiv:2509.08827, 2025.

\bibitem{wang2024survey}
L. Wang, C. Ma, X. Feng, Z. Zhang, H. Yang, J. Zhang {\it et al.}, ``A survey on large language model based autonomous agents,'' {\it Front. Comput. Sci.}, vol. 18, no. 6, p. 186345, 2024.

\bibitem{li2024survey}
X. Li, S. Wang, S. Zeng, Y. Wu, and Y. Yang, ``A survey on {LLM}-based multi-agent systems: Workflow, infrastructure, and challenges,'' {\it Vicinagearth}, vol. 1, no. 1, p. 9, 2024.

\bibitem{zhao2025llm}
B. Zhao, L. G. Foo, P. Hu, C. Theobalt, H. Rahmani, and J. Liu, ``{LLM}-based agentic reasoning frameworks: A survey from methods to scenarios,'' arXiv:2508.17692, 2025.

\bibitem{tran2025multi}
K. T. Tran, D. Dao, M.-D. Nguyen, Q.-V. Pham, B. O'Sullivan, and H. D. Nguyen, ``Multi-agent collaboration mechanisms: A survey of {LLMs},'' arXiv:2501.06322, 2025.

\bibitem{rasal2024navigating}
S. Rasal and E. J. Hauer, ``Navigating complexity: Orchestrated problem solving with multi-agent {LLMs},'' arXiv:2402.16713, 2024.

\bibitem{du2025multi}
Z. Du, C. Qian, W. Liu, Z. Xie, Y. Wang, R. Qiu {\it et al.}, ``Multi-agent collaboration via cross-team orchestration,'' in {\it Proc. Findings Assoc. Comput. Linguistics}. ACL, 2025, pp. 10\,386--10\,406.

\bibitem{dang2025multi}
Y. Dang, C. Qian, X. Luo, J. Fan, Z. Xie, R. Shi {\it et al.}, ``Multi-agent collaboration via evolving orchestration,'' arXiv:2505.19591, 2025.

\bibitem{zhang2025osc}
J. Zhang, Y. Fan, K. Cai, X. Sun, and K. Wang, ``{OSC}: Cognitive orchestration through dynamic knowledge alignment in multi-agent {LLM} collaboration,'' arXiv:2509.04876, 2025.

\bibitem{LangGraph2024}
LangChain AI, ``{LangGraph},'' 2024. [Online]. Available: https://github.com/langchain-ai/langgraph

\bibitem{Liu_LlamaIndex_2022}
J. Liu, ``{LlamaIndex},'' 2022. [Online]. Available: https://github.com/run-llama/llama\_index

\bibitem{pmlr-v202-radford23a}
A. Radford, J. W. Kim, T. Xu, G. Brockman, C. Mcleavey, and I. Sutskever, ``Robust speech recognition via large-scale weak supervision,'' in {\it Proc. 40th Int. Conf. Mach. Learn.}, vol. 202. PMLR, 2023, pp. 28\,492--28\,518.

\bibitem{ravanelli2021speechbrain}
M. Ravanelli, T. Parcollet, P. Plantinga, A. Rouhe, S. Cornell, L. Lugosch {\it et al.}, ``{SpeechBrain}: A general-purpose speech toolkit,'' arXiv:2106.04624, 2021.

\bibitem{microsoft2025phi4}
A. Abouelenin, A. Ashfaq, A. Atkinson, H. Awadalla, N. Bach, J. Bao {\it et al.}, ``{Phi-4-Mini} technical report: Compact yet powerful multimodal language models via mixture-of-{LoRAs},'' arXiv:2503.01743, 2025.

\end{thebibliography}
\end{document}